\documentclass[twocolumn]{article}
\usepackage[utf8]{inputenc}
\usepackage{abstract}
\usepackage{graphicx}
\usepackage{amsmath}
\usepackage{amssymb}
\usepackage{stfloats}
\usepackage{multirow}
\usepackage{bm}
\usepackage{hyperref}
\hypersetup{colorlinks,allcolors=[rgb]{0,0.44,0.74}}
\usepackage[superscript]{cite}
\usepackage{geometry}
\usepackage{authblk}
\usepackage[font=small,labelfont=bf]{caption}
\usepackage{libertine}
\usepackage[T1]{fontenc}
\usepackage{inconsolata}   

\usepackage{xcolor,soul}    
\usepackage[most]{tcolorbox}

\definecolor{light-gray}{gray}{0.9} 
\definecolor{yellow}{RGB}{255, 233, 138} 

\tcbset{highlight math style={
  enhanced jigsaw,
  size=tight,
  colback=light-gray,
  boxrule=0pt,
  rounded corners,
  left=2pt,  
  right=2pt, 
  top=2pt,   
  bottom=2pt, 
  colframe=white,
  arc=3pt
}}

\newcommand{\code}[1]{\tcbhighmath{\small{\texttt{#1}}}}

\newcommand{\myhl}[1]{\tcbhighmath[colback=yellow,arc=2pt]{\text{#1}}}

\makeatletter  \renewcommand{\@biblabel}[1]{\textbf{#1}}  \makeatother

\title{Methods to integrate multinormals and compute classification measures}

\author[1,3,4]{Abhranil Das \thanks{abhranil.das@utexas.edu}}
\author[2,3,4]{Wilson S Geisler}
\affil[1]{Department of Physics, The University of Texas at Austin}
\affil[2]{Department of Psychology, The University of Texas at Austin}
\affil[3]{Center for Perceptual Systems, The University of Texas at Austin}
\affil[4]{Center for Theoretical and Computational Neuroscience, The University of Texas at Austin}

\date{Published: Dec 23, 2020. Last revised: \today. \\
\ \\
\myhl{Note:} This arxiv version has several corrections and extensions over the Journal of Vision version `A method to integrate and classify normal distributions'. \cite{das2021method}}

\begin{document}

\maketitle

\begin{abstract}
    Univariate and multivariate normal probability distributions are widely used when modeling decisions under uncertainty. Computing the performance of such models requires integrating these distributions over specific domains, which can vary widely across models. Besides some special cases, there exist no general analytical expressions, standard numerical methods or software for these integrals. Here we present mathematical results and open-source software that provide (i) the probability in any domain of a normal in any dimensions with any parameters, (ii) the probability density, cumulative distribution, and inverse cumulative distribution of any function of a normal vector, (iii) the classification errors among any number of normal distributions, the Bayes-optimal discriminability index and relation to the operating characteristic, (iv) ways to scale the discriminability of two distributions, (v) dimension reduction and visualizations for such problems, and (vi) tests for how reliably these methods may be used on given data. We demonstrate these tools with vision research applications of detecting occluding objects in natural scenes, and detecting camouflage.
    
    \medskip
    \noindent \textbf{Keywords:} multivariate normal, integration,  classification, signal detection theory, Bayesian ideal observer, vision 
\end{abstract}

\section{Introduction}
The univariate or multivariate normal (henceforth called simply `normal') is arguably the most important and widely‐used probability distribution. It is frequently used because various central‐limit theorems guarantee that normal distributions will occur commonly in nature, and because it is the simplest and most tractable distribution that allows
arbitrary correlations between the variables.

Normal distributions form the basis of many theories and models in the natural and social sciences. For example, they are the foundation of Bayesian statistical decision/classification theories using Gaussian discriminant analysis \cite{ng2000cs229}, and are widely applied in diverse fields such as vision science, neuroscience, probabilistic planning in robotics, psychology, and economics. These theories specify optimal performance under uncertainty, and are often used to provide a benchmark against which to evaluate the performance (behavior) of humans, other animals, neural circuits or algorithms. They also serve as a starting point in developing other models/theories that describe sub‐optimal performance of agents.

To compute the performance predicted by such theories it is necessary to integrate the normal distributions over specific domains. For example, a particularly common task in vision science is classification into two categories (e.g. detection and discrimination tasks). The predicted maximum accuracy in such tasks is determined by integrating normals over domains defined by a quadratic decision boundary \cite{green1966signal,duda2012pattern}. Predicted accuracy of some of the possible sub‐optimal models is determined by integrating over other domains.

Except for some special cases \cite{ruben1960probability,genz2002comparison,genz2009computation} (e.g. multinormal probabilities in rectangular domains, or flat domains such as when two normals have equal covariance), there exists no general analytical expression for these integrals, and we must use numerical methods, such as integrating over a grid, or Monte-Carlo integration.

Since the normal distribution tails off infinitely outwards, it is inefficient to numerically integrate it over a finite uniform Cartesian grid, which would be large and collect ever-reducing masses outwards, yet omit some mass wherever the grid ends. Also, if the normal is elongated by unequal variances and strong covariances, or the integration domain is complex and non-contiguous, na\"ive integration grids will waste resources in regions or directions that have low density or are outside the domain. One then needs to visually inspect and arduously hand-tailor the integration grid to fit the  shape of each separate problem.

Monte-Carlo integration involves sampling from the multinormal, then counting the fraction of samples in the integration domain. This does not have the above inefficiencies or problems, but has other issues. Unlike grid integration, the desired precision cannot be specified, but must be determined by measuring the spread across multiple runs. Also, when the probability in the integration domain is very small (e.g. to compute the classification error rate or discriminability $d'$ for highly-separated normal distributions), it cannot be reliably sampled without a large number of samples, which costs resource and time (see the performance benchmark section for a comparison).

Thus, there is no single analytical expression, numerical method or standard software tool to quickly and accurately integrate arbitrary normals over arbitrary domains, or to compute classification errors and the discriminability index $d’$. Evaluating these quantities is often simplified by making the limiting assumption of equal variance. This impedes the quick testing, comparison, and optimization of models. Here we describe a mathematical method and accompanying software implementation which provides functions to (i) integrate normals with arbitrary means and covariances in any number of dimensions over arbitrary domains, (ii) compute the pdf, cdf and inverse cdf of any function of a multinormal variable (normal vector), and (iii) compute the
performance of classifying amongst any number of normals. This software is available as a Matlab toolbox \href{https://www.mathworks.com/matlabcentral/fileexchange/84973-integrate-and-classify-normal-distributions}{`Integrate and classify normal distributions'}, and the source code is at \href{https://github.com/abhranildas/IntClassNorm}{github.com/abhranildas/IntClassNorm}.

We first review and assimilate previous mathematical results into a generalized chi-squared method that can integrate arbitrary normals over quadratic domains. Then we present our ray-trace method to integrate arbitrary normals over any domain, and consequently to compute the distribution of any real‐valued function of a normal vector. We describe how these results can be used to compute error rates (and other relevant quantities) for Bayes-optimal
and custom classifiers, given arbitrary priors and outcome cost matrix. We then present some methods to reduce problems to fewer dimensions, for analysis or visualization. Next, we provide a way to test whether directly measured samples from the actual distributions in a classification problem are close
enough to normal to trust the computations from the toolbox. After describing the methods and software toolbox with examples, we demonstrate their accuracy and speed across a variety of problems. We show that for quadratic-domain problems both the generalized chi-squared method and the ray-trace method are accurate, but vary in relative speed depending on the particular problem. Of course, for domains that are not quadratic, only the ray-trace method applies. Finally, we illustrate the methods with two applications from our laboratory: modeling detection of occluding targets in natural scenes, and detecting camouflage.

\section{Integrating the normal}
\subsection{In quadratic domains: the generalized chi-square method}
Integrating the normal in quadratic domains is important for computing the maximum possible classification accuracy. The problem is the following: given a column vector $\bm{x} \sim N(\bm{\mu}, \bm{\Sigma}$), find the probability that 
\begin{equation}
    q(\bm{x})=\bm{x}' \mathbf{Q}_2 \bm{x} + \bm{q}_1' \bm{x} + q_0 > 0.
\end{equation}
(Here and henceforth, bold upper-case symbols represent matrices, bold lower-case symbols represent vectors, and regular lower-case symbols represent scalars.)

This can be viewed as the multi-dimensional integral of the normal probability over the domain $q(\bm{x}) > 0$ (that we call the `normal probability view'), or the single-dimensional integral of the probability of the scalar quadratic function $q(\bm{x})$ of a normal vector, above 0 (the `function probability view').

We may consider this distribution as a linear transformation of a unit spherical multinormal: $\bm{x}=\mathbf{S}\bm{z}+\bm{\mu}$, where $\bm{z}$ is standard multinormal and $\mathbf{S}=\bm{\Sigma}^\frac{1}{2}$ is the symmetric square root, which can be called the `standard deviation matrix' and regarded as the multi-dimensional sd, since it linearly scales the normal, like $\sigma$ does in 1d. Its eigenvectors and values are the axes of the 1 sd error ellipsoid.

We now invert the linear transform to standardize the normal: $\bm{z}=\mathbf{S}^{-1} (\bm{x}-\bm{\mu})$. This is a Mahalanobis whitening transform \cite{kessy2018optimal}, and it transforms the integration domain to a different quadratic:
\begin{align} \label{eq:quadratic_domain}
    \begin{split}
        \tilde{q}(\bm{z}) &= \bm{z}' \mathbf{\widetilde{Q}}_2 \bm{z} + \tilde{\bm{q}}_1' \bm{z} + \tilde{q}_0 > 0, \text{with} \\
        \mathbf{\widetilde{Q}}_2 &= \mathbf{S} \mathbf{Q}_2 \mathbf{S}, \\
        \tilde{\bm{q}}_1 &= 2\mathbf{S}\mathbf{Q}_2 \bm{\mu}+\mathbf{S}\bm{q}_1, \\
        \tilde{q}_0 &= q(\bm{\mu}).
    \end{split}
\end{align}
Now the problem is to find the probability of the standard normal $\bm{z}$ in this domain. If there is no quadratic term $\mathbf{\widetilde{Q}}_2$, $\tilde{q}(\bm{z})$ is normally distributed, the integration domain boundary is a flat, and the probability is $\Phi(\frac{\tilde{q}_0}{\lVert \tilde{\bm{q}}_1 \rVert})$, where $\Phi$ is the standard normal cdf \cite{ruben1960probability}. Otherwise, say $\mathbf{\widetilde{Q}}_2=\mathbf{RDR}'$ is its eigen-decomposition, where $\mathbf{R}$ is orthogonal, i.e. a rotoreflection. So $\bm{y}=\mathbf{R}'\bm{z}$ is also standard normal, and in this space the quadratic is:
\begin{align*}
    \hat{q}(\bm{y}) &=\bm{y}' \mathbf{D}\bm{y} + \bm{b}' \bm{y} + \tilde{q}_0 \quad \quad \left(\bm{b}=\mathbf{R}'\tilde{\bm{q}}_1\right) \\
    &=\sum_i \left(D_i y_i^2 + b_i y_i \right) + \sum_{i'} b_{i'} y_{i'} + \tilde{q}_0 \\
    &\text{($i$ and $i'$ index the nonzero and zero eigenvalues)} \\
    &=\sum_i D_i \left(y_i + \frac{b_i}{2D_i} \right)^2 + \sum_{i'} b_{i'} y_{i'} + \tilde{q}_0 - \sum_i D_i \left(\frac{b_i}{2D_i} \right)^2 \\
    &=\sum_i D_i \, \chi'^2_{1,(b_i/2D_i)^2} + x,
\end{align*}
a weighted sum of non-central chi-square variables $\chi'^2$, each with 1 degree of freedom, and a normal variable $x \sim N(m,s^2)$. So this is a generalized chi-square variable $\tilde{\chi}_{\bm{w}, \bm{k}, \bm{\lambda},s,m}$, where we merge the non-central chi-squares with the same weights, so that the vector of their weights $\bm{w}$ are the \textit{unique} nonzero eigenvalues among $D_i$, their degrees of freedom $\bm{k}$ are the numbers of times the eigenvalues occur, and their non-centralities, and normal parameters are:
\begin{equation*}
    \lambda_j =\frac{1}{4 w_j^2} \sum_{i: D_i=w_j} b_i^2, \quad s = \sqrt{\sum_{i'} b_{i'}^2}, \quad m =q(\bm{\mu})- \bm{w.\lambda}.
\end{equation*}

In a later paper \cite{das2024new}, we derive the inverse map, i.e. from the parameters of a generalized chi-square distribution, to the parameters of the corresponding quadratic form of a multinormal.

The required probability, $p \left(\tilde{\chi} > 0 \right)$, is now a 1d integral, computable using, say, Ruben's \cite{ruben1962probability}, Imhof's \cite{imhof1961computing,davies1973numerical}, IFFT \cite{das2024new} or ray \cite{das2024new} methods. We use the Matlab toolbox \href{https://www.mathworks.com/matlabcentral/fileexchange/85028-generalized-chi-square-distribution}{`Generalized chi-square distribution'} that we developed (source code is at \href{https://github.com/abhranildas/gx2-matlab}{github.com/abhranildas/gx2-matlab}), which can compute the generalized chi-square parameters corresponding to a quadratic form of a normal vector, its statistics, cdf (using three different methods), pdf, inverse cdf and random numbers. We have also ported this to a \href{https://pypi.org/project/gx2/}{`gx2' python package} with the same functionality (source code is at \href{https://github.com/abhranildas/gx2-py}{github.com/abhranildas/gx2-py}).

Previous software implements specific forms of this theory for particular quadratics such as ellipsoids \cite{genz2009computation}. The method described here correctly handles all quadratics (ellipsoids, hyperboloids, paraboloids and degenerate conics) in all dimensions.

\subsection{In any domain: the ray-trace method}

We present below our method to integrate the normal distribution in an arbitrary domain, which takes an entirely different approach than the generalized chi-square method. Genz and Bretz \cite{genz2002comparison} outlined this approach and mentioned that it may be used for general domains, but the method they explicitly developed was for (hyper-)rectangular domains only. Here we explicitly develop the method in detail for general domains. A broad overview is as follows. We first standardize the normal to make it spherically symmetric, then we integrate it in spherical polar coordinates, outwards from the center. We first calculate the radial integral by sending `rays' from the center to trace out the integration domain in every direction, i.e. determine the points where each ray crosses into and out of the domain (akin to the computer graphics method of ray-tracing, that traces light rays outward from the projection center to compute where it hits the different objects it has to render). By knowing these crossing points, we then calculate the probability amount on each ray. Then we add up these probabilities over all the angles. This method of breaking up the problem produces fast and accurate results to arbitrary tolerance for all problem shapes, without needing any manual adjustment.

\subsubsection{Standard polar form}
The problem is to find the probability that $f(\bm{x})>0$, where $f(\bm{x})$ is a sufficiently general function (with a finite number of zeros in any direction within the integration span around the normal mean, i.e. without rare pathologies such as the Dirichlet function with infinite zeros in any interval). As before, we first standardize the space to obtain $\tilde{f}(\bm{z})=f(\mathbf{S} \bm{z}+\bm{\mu})$. Then we switch to polar axis-angle coordinates $z$ and $\bm{n}$: any point $\bm{z}=z\bm{n}$, where the unit vector $\bm{n}$ denotes the angle of that point, and $z$ is its coordinate along the axis in this direction. Then the integral can be written as:
\begin{equation*}
    \int_{\tilde{\Omega}} (2\pi)^{-\frac{k}{2}} e^{-\frac{z^2}{2}} d\bm{z} = \int_{\bm{n}} d\bm{n} \underbrace{\int_{\tilde{\Omega}_{\bm{n}}} (2\pi)^{-\frac{k}{2}} e^{-\frac{z^2}{2}} z^{k-1} dz}_{\text{axial integral}}.
\end{equation*}
where $\tilde{\Omega}$ is the domain where $\tilde{f}(\bm{z})>0$, and $\tilde{\Omega}_{\bm{n}}$ is its slice along the axis $\bm{n}$, i.e. the intervals along the axis where the axial domain function $\tilde{f}_{\bm{n}}(z)=\tilde{f}(z\bm{n})>0$. This may be called the `standard polar form' of the integral. $d\bm{n}$ is the differential angle element ($d\theta$ in 2d, $\sin\theta \,  d\theta \, d\phi$ in 3d, $\sin\theta \, \sin^2 \psi \, d\theta \, d\phi \, d\psi$ in 4d etc).

\begin{figure}[t]
    \includegraphics[width=\columnwidth]{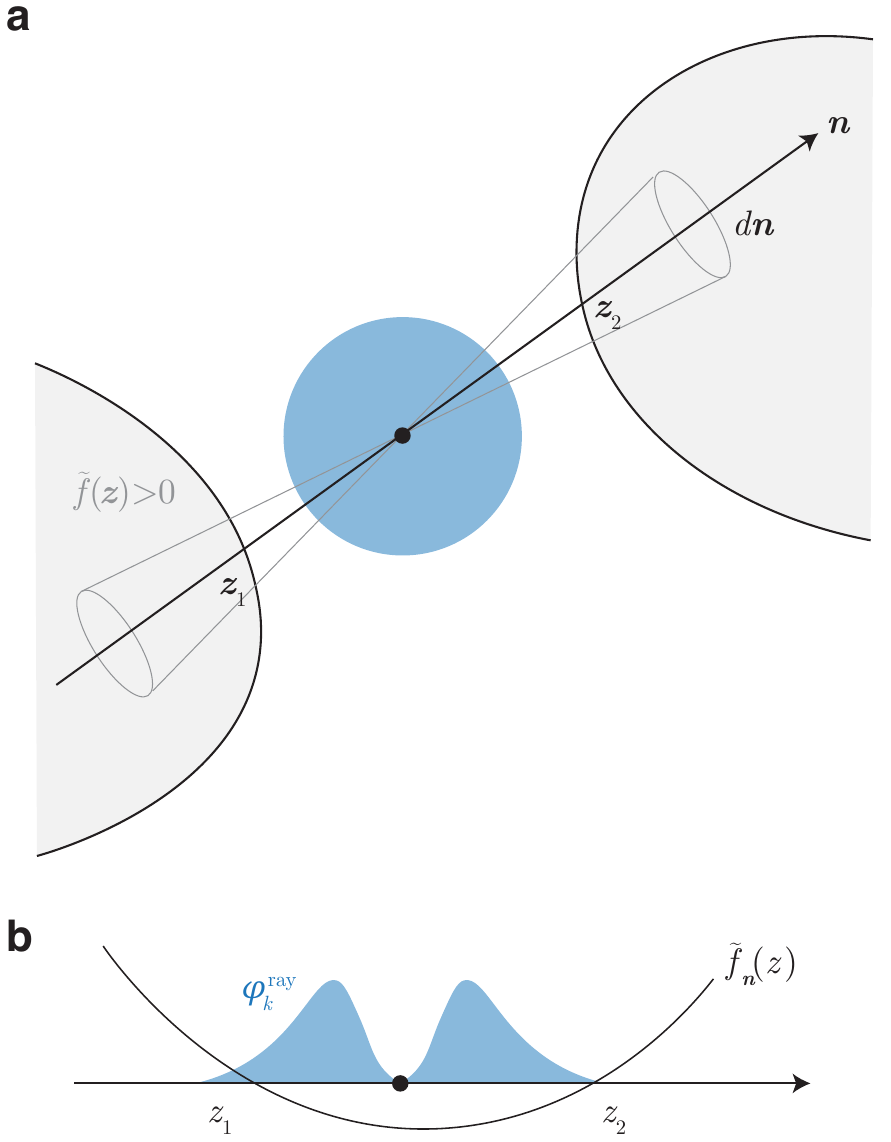}
    \caption{Method schematic. \textbf{a.} Standard normal error ellipse is blue. Arrow indicates a ray from it at angle $\bm{n}$ in an angular slice $d\bm{n}$, crossing the gray integration domain $\tilde{f}(\bm{z})>0$ at $\bm{z}_1$ and $\bm{z}_2$. \textbf{b.} 1d slice of this picture along the ray. The standard normal density along a ray is blue. $\tilde{f}_{\bm{n}}(z)$ is the slice of the domain function $\tilde{f}(\bm{z})$ along the ray, crossing 0 at $z_1$ and $z_2$.}
    \label{fig:1_schematic}
\end{figure}

\subsubsection{Integration domain on a ray}
First let us consider the axial integration along direction $\bm{n}$. Imagine that we `trace' the integration domain with an axis through the origin in this direction (a bi-directional `ray' marked by the arrow in fig. \ref{fig:1_schematic}a), i.e. determine the part of this ray axis that is in the integration domain, defined by $\tilde{f}_{\bm{n}}(z)>0$. For example, if the integration domain is a quadratic such as eq. \ref{eq:quadratic_domain}, its 1d trace by the ray is given by:
\begin{align*}
    \tilde{q}_{\bm{n}}(z)=\tilde{q}(z\bm{n})&=\bm{n}' \mathbf{\widetilde{Q}}_2 \bm{n} z^2 +\tilde{\bm{q}}_1' \bm{n}z + \tilde{q}_0 \\
    &=\tilde{q}_2(\bm{n}) \, z^2 + \tilde{q}_1(\bm{n}) \, z + \tilde{q}_0>0.
\end{align*}
This is a scalar quadratic domain in $z$ that varies with the direction. Fig. \ref{fig:1_schematic}b is an example of such a domain. The ray domain function $\tilde{f}_{\bm{n}}$ crosses 0 at $z_1$ and $z_2$, and the integration domain is below $z_1$ (which is negative), and above $z_2$.

Note that a sufficient description of such domains on an axis is to specify all the points at which the domain function crosses zero, and its overall sign, which determines which regions are within and which are outside the domain (so any overall scaling of the domain function does not matter). That is, we specify whether or not the beginning of the ray (at $-\infty$) is inside the domain, and all the points at which the ray crosses the domain. We denote the first by the initial sign $\psi(\bm{n}) = \text{sign} (\tilde{f}_{\bm{n}}(-\infty)) =1/{-}1/0$ if the ray begins inside/outside/grazing the integration domain. For a quadratic domain, for example:
\begin{equation*}
    \psi(\bm{n}) = \text{sign} \left(\tilde{q}_{\bm{n}} \left(-\infty \right) \right) 
    = \begin{cases}
    \text{sign} \left(\tilde{q}_2 \left(\bm{n} \right)\right), \text{  if } \tilde{q}_2 (\bm{n})\neq 0, \\
    - \text{sign} \left(\tilde{q}_1 \left(\bm{n} \right)\right), \text{  if } \tilde{q}_2 (\bm{n}) = 0, \\
    \text{sign} \left(\tilde{q}_0 \right), \text{ if } \tilde{q}_2 (\bm{n}) = \tilde{q}_1 (\bm{n}) = 0.
    \end{cases}
\end{equation*}

We can encapsulate all three cases into the simple clever expression:
\begin{equation*}
\psi(\bm{n})= \text{sign}\left[ 4 \text{ sign}\left(\tilde{q}_2 \left(\bm{n} \right)\right) - 2 \text{ sign} \left(\tilde{q}_1 \left(\bm{n} \right)\right) + \text{sign}\left(\tilde{q}_0 \right) \right].
\end{equation*}
The crossing points are the zeros $z_i(\bm{n})$ of $\tilde{f}_{\bm{n}}(z)=f(z\mathbf{S} \bm{n}+\bm{\mu})$ ($z_i\bm{n}$ are then the boundary points in the full space). For a quadratic domain $\tilde{q}_{\bm{n}}(z)$, these are simply its roots. For a general domain, the zeros are harder to compute. Chebyshev polynomial approximations \cite{trefethen2019approximation} aim to find all zeros of a general function, but can be slow. Other numerical algorithms can find all function zeros in an interval to arbitrary accuracy. We use such an algorithm to find the zeros of $\tilde{f}_{\bm{n}}(z)$ within $(-m,m)$. This amounts to ray-tracing $f(\bm{x})$ within a Mahalanobis distance $m$ of the normal. The maximum error in the integral due to this approximation is the standard multinormal probability beyond a radius of $m$, which in $k$ dimensions is $\bar{F}_{\chi_k}(m)$, the complementary cdf of the chi distribution with $k$ degrees of freedom.

In fig. \ref{fig:1_schematic}, the initial sign along the ray is 1, and $z_1$ and $z_2$ are the crossing points.

Most generally, this method can integrate in any domain for which we can return its `trace' (i.e. the initial sign and crossing points) along any ray $\bm{n}$ through any origin $\bm{o}$. So if a domain is already supplied in the form of these `ray-trace' functions $\psi(\bm{o},\bm{n})$ and $z_i(\bm{o},\bm{n})$, our method can readily integrate over it. For example, the ray-trace function of the domain $y>k$ in 2d returns $\psi=-\text{sign} (n_y)$ and $z= \frac{k-o_y}{n_y}$. When supplied with quadratic domain coefficients, or a general implicit domain $f(\bm{x})>0$, the toolbox ray-traces it automatically under the hood. For an implicit domain, the numerical root-finding works only in a finite interval, and is slower and may introduce small errors. So, if possible, a slightly faster and more accurate alternative to the implicit domain format is to directly construct its ray-trace function by hand.

\begin{figure*}[!h] \includegraphics[width=\textwidth]{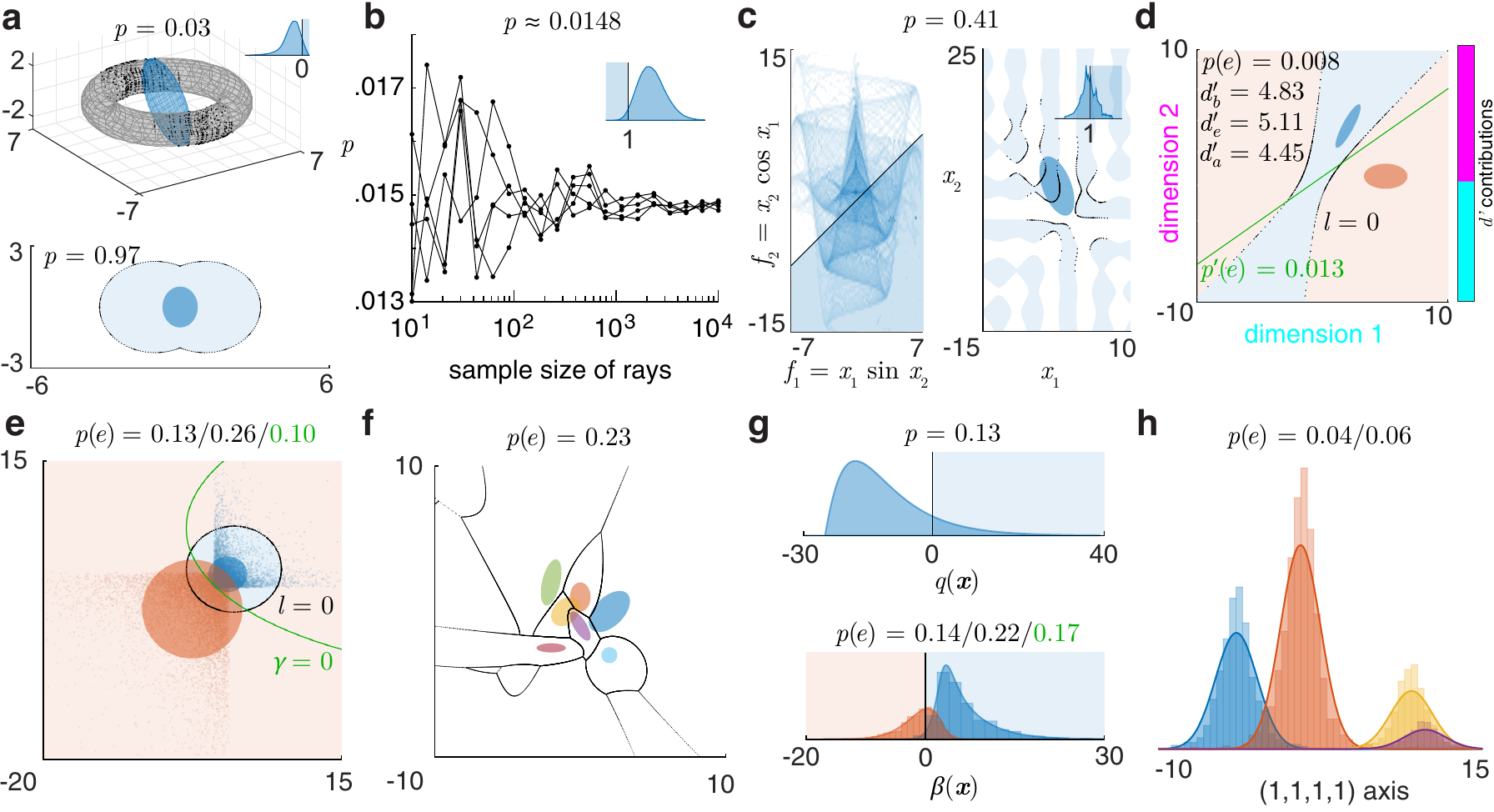}
    \caption{Toolbox outputs for some integration and classification problems. \textbf{a.} Top: the probability of a 3d normal (blue shows 1 sd error ellipsoid) in an implicit toroidal domain $f_t(\bm{x})>0$. Black dots are boundary points within 3 sd traced by the ray method, across Matlab's adaptive integration grid over angles. Inset: pdf of $f_t(\bm{x})$ and its integrated part (blue overlay). Bottom: integrating a 2d normal (blue error ellipse) in a domain built by the union of two circles. \textbf{b.} Estimates of the 4d standard normal probability in the 4d polyhedral domain $f_p(\bm{x})= \sum_{i=1}^4 \lvert x_i \rvert<1$ using the ray-trace method with Monte Carlo ray-sampling, across 5 runs, converging with growing sample size of rays. Inset: pdf of $f_p(\bm{x})$ and its integrated part. \textbf{c.} Left: heat map of joint pdf of two functions of a 2d normal, to be integrated over the implicit domain $f_1-f_2>1$ (overlay). Right: corresponding integral of the normal over the domain $h(\bm{x})=x_1 \sin x_2 - x_2 \cos x_1 >1$ (blue regions), `traced' up to 3 sd (black dots). Inset: pdf of $h(\bm{x})$ and its integrated part. \textbf{d.} Classifying two 2d normals using the optimal boundary $l$, which yields the Bayes-optimal discriminability $d'_b$. Color-bar shows the proportions by which the two dimensions contribute to the overall discriminability. $d'_e$ and $d'_a$ are approximate discriminability indices. A custom suboptimal boundary such as the green line can also be used for classification.  \textbf{e.} Classification based on samples (dots) from non-normal distributions. Filled ellipses are error ellipses of fitted normals. $\gamma$ is an optimized boundary between the samples. The three error rates are: of the normals with $l$, of the samples with $l$, and of the samples with $\gamma$. \textbf{f.} Classifying several 2d normals with arbitrary means and covariances. \textbf{g.} Top: 1d projection of a 4d normal integral over a quadratic domain $q(\bm{x})>0$. Bottom: Projection of the classification of two 4d normals based on samples, with unequal priors, and unequal outcome values (correctly classifying the blue class is valued 4 times the red, hence the optimal criterion is shifted), onto the axis of the Bayes decision variable $\beta$. Histograms and smooth curves are the projections of the samples and the fitted normals. The sample-optimized boundary $\gamma=0$ cannot be uniquely projected to this $\beta$ axis. \textbf{h.} Classification based on four 4d non-normal samples, with different priors and outcome values, projected on the axis along (1,1,1,1). The boundaries cannot be projected to this axis.}
    \label{fig:2_examples}
\end{figure*}

\subsubsection{Standard normal distribution on a ray}
In order to integrate over piecewise intervals of $z$ such as fig. \ref{fig:1_schematic}b, we shall first calculate the semi-definite integral up to some $z$, then stitch them together over the intervals with the right signs.

Consider the probability in the angular slice $d\bm{n}$ below some negative $z$ such as $z_1$ in fig. \ref{fig:1_schematic}a. Note that the probability of a standard normal beyond some radius is given by the chi distribution. If $\Omega_k$ is the total angle in $k$ dimensions (2 in 1d, $2\pi$ in 2d, $4\pi$ in 3d, $2\pi^2$ in 4d), and $F_{\chi_k}(x)$ is the cdf of the chi distribution with $k$ degrees of freedom, we have:
\begin{equation*}
    \Omega_k \int_{-\infty}^{z<0} (2\pi)^{-\frac{k}{2}} e^{-\frac{z^2}{2}} z^{k-1} dz = 1-F_{\chi_k}(\lvert z \rvert).
\end{equation*}
So the probability in the angular slice $d\bm{n}$ below a negative $z$ is $\left[1-F_{\chi_k}(\lvert z \rvert)\right]\frac{d\bm{n}}{\Omega_k}$.
Now, for the probability in the angular slice below a positive $z$ (such as $z_2$), we need to add two probabilities: that in the finite cone from the origin to the point, which is $F_{\chi_k}(z) \frac{d\bm{n}}{\Omega_k}$, and that in the entire semi-infinite cone on the negative side, which is $\frac{d\bm{n}}{\Omega_k}$, to obtain $\left[1+F_{\chi_k}(z) \right] \frac{d\bm{n}}{\Omega_k}$. Thus, the probability in an angular slice $d\bm{n}$ below a positive or negative $z$ is $\left[1+\text{sign}(z) F_{\chi_k}(\lvert z \rvert) \right]  \frac{d\bm{n}}{\Omega_k}$. We normalize this by the total probability in the angular slice, $2\frac{d\bm{n}}{\Omega_k}$, to define the distribution of the standard normal along a ray: $\Phi^{\text{ray}}_k(z)=\left[1+\text{sign}(z) F_{\chi_k}(\lvert z \rvert) \right]/2$. Its density is found by differentiating: $\phi^{\text{ray}}_k(z) = f_{\chi_k}(\lvert z \rvert)/2$, so it is simply the chi distribution symmetrically extended to negative numbers. Notice that $\phi^{\text{ray}}_1(z) = \phi(z)$, but in higher dimensions it rises, then falls outward (fig. \ref{fig:1_schematic}b), due to the opposing effects of the density falling but the volume of the angular slice growing outward. Since Matlab does not yet incorporate the chi distribution, we instead define, in terms of the chi-square distribution, $\Phi^{\text{ray}}_k(z) =  \left[1+\text{sign}(z) F_{\chi^2_k}(z^2) \right]/2$ and $\phi^{\text{ray}}_k(z) = \lvert z \rvert f_{\chi^2_k}(z^2)$.

\subsubsection{Probability in an angular slice}
We can now write the total probability in the angular slice of fig. \ref{fig:1_schematic} as the sum of terms accounting for the initial sign and each root. The total volume fraction of the double cone is $\frac{2 d\bm{n}}{\Omega_k}$. Now first consider only the initial sign and no roots. Then if the ray starts inside the domain ($\psi=1$), it stays inside, and the probability content is $\frac{2 d\bm{n}}{\Omega_k}$. If it begins and stays outside ($\psi=-1$), it is 0. And if it grazes the domain throughout ($\psi=0$), half of the angular volume is inside the domain and half is outside, so the probability is $\frac{d\bm{n}}{\Omega_k}$. So without accounting for roots, the probability in general is $\frac{\psi(\bm{n})+1}{\Omega_k}$. To this we add, sequentially for each root, the probability from the root to $\infty$, signed according to whether we are entering or exiting the domain at that root. So we have, for fig. \ref{fig:1_schematic},
\begin{equation*}
    dp \, (\bm{n}) = \left[\frac{2}{\Omega_k} - \frac{2\bar{\Phi}^{\text{ray}}_k(z_1)}{\Omega_k} + \frac{2\bar{\Phi}^{\text{ray}}_k(z_2)}{\Omega_k} \right] d\bm{n}.
\end{equation*}
The sign of the first root term is always opposite to $\psi$, and subsequent signs alternate as we enter and leave the domain. In general then, we can write:
\begin{align*}
    dp \, (\bm{n}) 
    &= \underbrace{\left[\psi(\bm{n})+1+2\psi(\bm{n}) \sum_i (-1)^i \, \bar{\Phi}^{\text{ray}}_k\left(z_i\left(\bm{n}\right)\right) \right]}_{\alpha(\bm{n})} \frac{d\bm{n}}{\Omega_k}
\end{align*}

Thus, the axial integral is $\frac{\alpha(\bm{n})}{\Omega_k}$. The total probability $ \frac{1}{\Omega_k} \int \alpha(\bm{n}) \  d\bm{n}$ can be computed, for up to 4d, by numerically integrating $\alpha(\bm{n})$ over a grid of angles spanning half the angular space (since we account for both directions of a ray), using any standard scheme. An adaptive grid can match the shape of the integration boundary (finer grid at angles where the boundary is sharply changing), and also set its fineness to evaluate the integral to a desired absolute or relative precision. Fig. \ref{fig:2_examples}a, top, illustrates integrating a trivariate normal with arbitrary covariance in an implictly-defined toroidal domain $f_t(\bm{x})= a - \left(b- \sqrt{x_1^2 + x_2^2} \right)^2-x_3^2 >0$.

Beyond 4d (or even in general), we can use Monte Carlo integration over the angles. We draw a sample of random numbers from the standard multinormal in those dimensions, then normalize their magnitudes, to get a uniform random sample of rays $\bm{n}$, over which the expectation $\langle \alpha(\bm{n}) \rangle/2$ is the probability estimate. Fig. \ref{fig:2_examples}b shows the computation of the 4d standard normal probability in the domain $ f_p(\bm{x}) = \sum_{i=1}^4 \lvert x_i \rvert < 1$, a 4d extension of a regular octahedron with plane faces meeting at sharp edges. This Monte-Carlo integration can be done in parallel. So, just like ray-tracing for graphics rendering, this method can run on the GPU, and automatically uses one when available, batching the rays to prevent memory overflow. This provides up to a 4x speedup.

Since the algorithm already computes the boundary points over its angular integration grid, they may be stored for plotting and inspecting the boundary. Rather than an adaptive integration grid though, boundaries are often best visualized over a uniform grid (uniform array of angles in 2D, or a Fibonacci sphere in 3D \cite{saff1997distributing}), which we can explicitly supply for this purpose.

\subsubsection{Set operations on domains}
Some applications require more complex integration or classification domains built using set operations (inversion/union/intersection) on simpler domains. With implicit domain formats this is easy. For example, if $f_A(\bm{x})>0$ and $f_B(\bm{x})>0$ define two domains $A$ and $B$, then $A^c$, $A \cap B$, and $A \cup B^c$ are described by $-f_A(\bm{x})>0$, $\min (f_A(\bm{x}),f_B(\bm{x}))>0$ and $\max (f_A(\bm{x}),-f_B(\bm{x}))>0$ respectively. Fig. \ref{fig:2_examples}a, bottom, illustrates integrating a 2d normal in a domain built by the union of two circles.

As we noted before, computations are faster and more accurate when domains are supplied in explicit ray-trace form, than as implicit functions. The toolbox provides functions to convert quadratic and general implicit domains to ray-trace format, and functions to use set operations on these to build complex ray-trace domains. For example, when a domain is inverted, only the initial sign of a ray through it flips, and for the intersection of several domains, the initial sign of a ray is the minimum of its individual initial signs, and the roots are found by collecting those roots of each domain where every other domain is positive.

\subsubsection{Probabilities of functions of a normal vector}

We previously mentioned the equivalent `normal probability' and `function probability' views of conceptualizing a normal integral. So far we have mostly used the normal probability view, seeing scalar functions $f(\bm{x})$ as defining integral domains of the normal $\bm{x}$. But in the function probability view, $f(\bm{x})$ is instead seen as a mapping from the multi-dimensional variable $\bm{x}$ to a scalar, which in some contexts can be considered a decision variable. Hence, integrating the normal in the multi-dimensional domain $f(\bm{x})>0$ corresponds to integrating the 1d pdf of the decision variable $f(\bm{x})$ beyond 0. It is sometimes helpful to plot this 1d pdf, especially when there are too many dimensions of $\bm{x}$ to visualize the normal probability view.

Conversely, given any scalar function $f(\bm{x})$ of a normal, its cdf,  $F_f(c)=p(f(\bm{x})<c)$, is computed as the normal probability in the domain $c-f(\bm{x})>0$. From this, we can obtain the pdf by numerically differentiating the cdf. In a later paper, we also analytically differentiate the ray-tracing method to directly compute the pdf of a scalar function of a normal vector. \cite{das2024new} (If it is a quadratic function, its generalized chi-square pdf can also be computed by convolving the constituent noncentral chi-square pdf's.) Figs. \ref{fig:2_examples}a-c and g show 1d pdf's of functions computed in this way. Also, inverting the function cdf using a numerical root-finding method gives us its inverse cdf.

With these methods to obtain the pdf, cdf and inverse cdf of functions of a normal vector, we can conveniently compute certain quantities. For example, if $x$ and $y$ are jointly normal with $\mu_x=1$, $\mu_y=2$, $\sigma_x=.1$, $\sigma_y=.2$, and $\rho_{xy}=.8$, we can compute the pdf, cdf and inverse cdf of the function $x^y$, and determine, say, that its mean, median and sd are respectively 1.03, 1 and 0.21.

The probability of a vector (multi-valued) function of the normal, e.g. $\bm{f}(\bm{x})=\begin{bmatrix} f_1(\bm{x}) & f_2(\bm{x}) \end{bmatrix}$, in some $\bm{f}$-domain (which may also be seen as the joint probability of two scalar functions), is again the normal probability in a corresponding $\bm{x}$-domain. For example, the joint cdf $F_{\bm{f}}(c_1,c_2)$ is the function probability in an explicit domain: $p\left(f_1<c_1,f_2<c_2\right)$, and can be computed as the normal probability in the intersection of the $\bm{x}$-domains $f_1(\bm{x})<c_1$ and $f_2(\bm{x})<c_2$, i.e. the domain $\min\left(c_1-f_1\left(\bm{x}\right),c_2-f_2\left(\bm{x}\right)\right)>0$. Numerically computing $\frac{\partial}{\partial c_1} \frac{\partial}{\partial c_2} F_{\bm{f}}(c_1,c_2)$ then gives the joint pdf of the vector function. Fig. \ref{fig:2_examples}c, left, is an example of a joint pdf of two functions of a bivariate normal with $\bm{\mu}=\begin{bmatrix} -2 & 5 \end{bmatrix}$ and $\bm{\Sigma} = \begin{bmatrix}
10 & -7 \\
-7 & 10
\end{bmatrix}$, computed in this way.

The probability of such a vector function in an implicit domain, i.e. $p\left(g\left(\bm{f}\right)>0\right)$, is computed as the normal probability in the implicit domain: $p\left(h\left(\bm{x}\right)>0 \right)$ where $h=g \circ \bm{f}$. Fig. \ref{fig:2_examples}c illustrates the function probability and normal probability views of the implicit integral $p(h=x_1 \sin x_2 - x_2 \cos x_1 >1)$. The 83rd percentile of this function $h$ (using the inverse cdf) is 4.87.

\section{Classifying normal samples}
Suppose observations come from several normal distributions with parameters $\bm{\mu}_i, \bm{\Sigma}_i$ and priors $p_i$, and the outcome values (rewards and penalties) of classifying them are represented in a matrix $\mathbf{V}$: $v_{ij}$ is the value of classifying a sample from $i$ as $j$.

If the true class is $i$, selecting $i$ over others provides a relative value gain of $v_i := v_{ii} - \sum_{j \neq i} v_{ij}$. Given a sample $\bm{x}$, the expected value gain of deciding $i$ is therefore $\langle v(i|\bm{x}) \rangle = p(i|\bm{x}) v_i = p(\bm{x}|i) \, p_i v_i$. The Bayes-optimal decision is to assign each sample to the class that maximizes this expected value gain, or its log:
\begin{equation*}
    \ln \langle v(i|\bm{x}) \rangle = -\frac{1}{2} (\bm{x} - \bm{\mu}_i)' \bm{\Sigma}_i^{-1} (\bm{x} - \bm{\mu}_i) + \ln \frac{p_i v_i}{\sqrt{\lvert \bm{\Sigma}_i \rvert (2\pi)^k}}.
\end{equation*}

When the outcome value is simply the correctness of classification, $\mathbf{V}=\bm{1}$ (so each $v_i = 1$), then this quantity is the log posterior, $\ln  p(i|\bm{x})$, and when priors are also equal, it is the log likelihood.

\subsection{Two normals}

Suppose there are only two normal classes $a$ and $b$. The Bayes-optimal decision rule is to pick $A$ if (upper-case denotes the estimated classes):
\begin{align} \label{eq:bdry_quadratic_coeffs}
\begin{split}
    &\ln \frac{\langle v(A|\bm{x}) \rangle }{\langle v(B|\bm{x}) \rangle} = \beta(\bm{x}) = \bm{x}' \mathbf{Q}_2 \bm{x} + \bm{q}_1' \bm{x} + q_0 > 0 \text{, where:} \\
    \ \\
    \mathbf{Q}_2 &= \frac{1}{2} \left(\bm{\Sigma}_b^{-1} - \bm{\Sigma}_a^{-1} \right), \\
    \bm{q}_1 &= \bm{\Sigma}_a^{-1} \bm{\mu}_a
    - \bm{\Sigma}_b^{-1} \bm{\mu}_b, \\
    q_0 &= \frac{1}{2} \left(\bm{\mu}_b' \bm{\Sigma}_b^{-1} \bm{\mu}_b
    - \bm{\mu}_a' \bm{\Sigma}_a^{-1} \bm{\mu}_a
    + \ln \frac{\lvert\bm{\Sigma}_b \rvert }{\lvert\bm{\Sigma}_a\rvert} \right) + \ln \frac{p_a v_a}{p_b v_b}.
    \end{split}
\end{align}

This quadratic $\beta(\bm{x})$ is the Bayes classifier, or the Bayes decision variable that, when compared to 0, maximizes expected gain.
  
When $\mathbf{V}=\bm{1}$, the Bayes decision variable is the log posterior ratio, and this decision rule minimizes overall error. The error rates of different types, i.e. true and false positives and negatives, are then the probabilities of the normals on either side of the quadratic boundary $\beta(\bm{x})=0$. These probabilities can be computed entirely numerically using the ray-trace method, or we can first arrive at mathematical expressions using the generalized chi-square method (as follows), which are then numerically evaluated. The overall error $p(e)$ is the prior-weighted sum of the error rates of each normal.

Further, when priors are equal, the Bayes decision variable is the log likelihood ratio (of $a$ vs $b$), which can be called $l(\bm{x})$.

\subsubsection{Single-interval (yes/no) task}

Consider a yes/no task where the stimulus $x$ comes from one of two equally likely 1d normals $a$ and $b$ with means $\mu_a,\mu_b$ and sd's $\sigma_a >\sigma_b$ (fig. \ref{fig:3_binary_class}a). The optimal decision (eq. \ref{eq:bdry_quadratic_coeffs}) is to pick $a$ if the Bayes decision variable (log likelihood ratio of $a$ vs $b$) $l_{\frac{a}{b}}(x)>0$, i.e. if
\begin{equation*}
    \left( \frac{x-\mu_b}{\sigma_b} \right)^2 - \left( \frac{x-\mu_a}{\sigma_a} \right)^2 + 2\ln \frac{\sigma_b}{\sigma_a} >0.
\end{equation*}
$l_{\frac{a}{b}}(x)$ is a scaled and shifted 1 dof noncentral chi-square for each class (fig. \ref{fig:3_binary_class}b), and the Bayes error rates are:
\begin{equation}\label{eq:1d_class}
\begin{gathered}
    p\left(B | a\right)=p\left(\chi'^2_{1,\sigma^2_a\lambda}<\sigma^2_b c\right), \quad p\left(A | b\right)=p\left(\chi'^2_{1,\sigma^2_b\lambda}>\sigma^2_a c\right),\\
    \text{where } \lambda=\left(\frac{\mu_a-\mu_b}{\sigma^2_a-\sigma^2_b}\right)^2, \quad c=\lambda+\frac{2 \ln \frac{\sigma_a}{\sigma_b}}{\sigma^2_a-\sigma^2_b},
\end{gathered}
\end{equation}

and $p(e)$ is their average.

\subsubsection{Two-interval task}

\begin{figure}[!b]
    \includegraphics[width=\columnwidth]{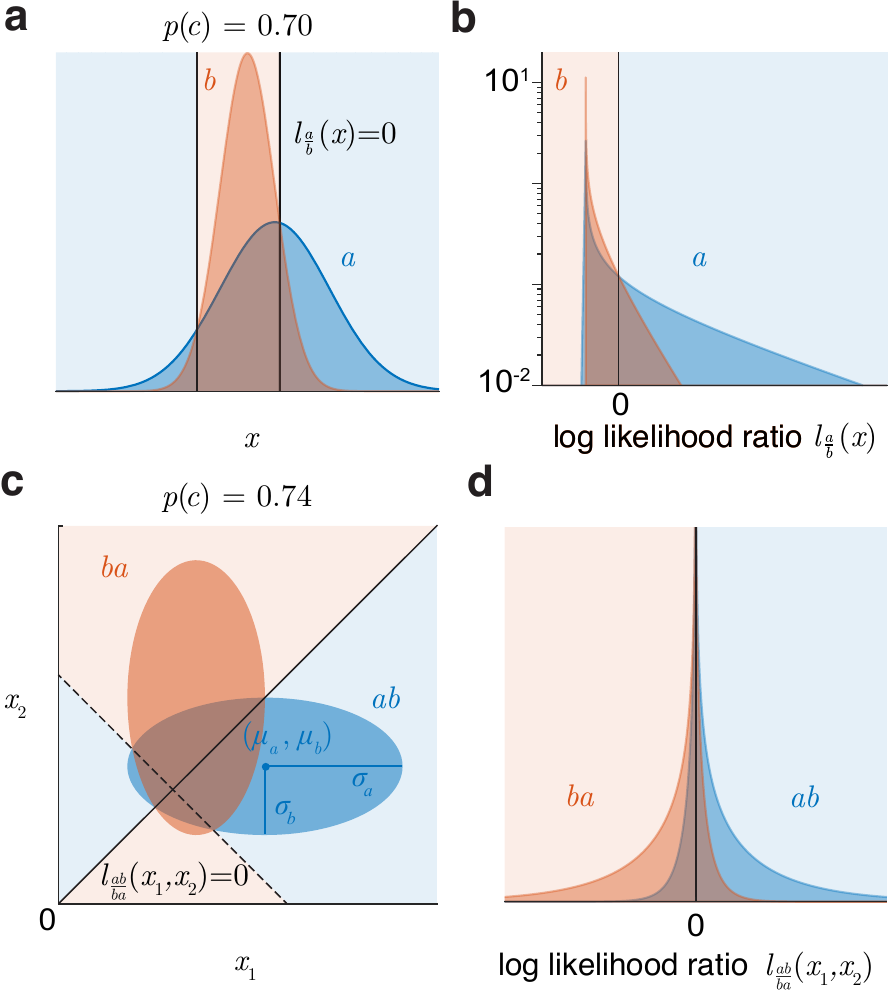}
    \caption{Binary yes/no and two-interval classification tasks. \textbf{a.} Optimal yes/no decision between two unequal-variance 1d normal distributions. \textbf{b.} The same task transformed to the log likelihood ratio axis (log vertical axis for clarity). \textbf{c.} Optimal two-interval discrimination between the same 1d normal distributions $a$ and $b$ is actually a discrimination between 2d normals $ab$ and $ba$. \textbf{d.} The task transformed to the log likelihood ratio axis.}
    \label{fig:3_binary_class}
\end{figure}

Now consider an equal-priors two-interval task, where two stimuli $x_1$ and $x_2$ come from each of the (general) distributions $a$ and $b$. A decision rule commonly employed here is to check which stimulus is larger \cite{simpson1973best,green2020homily}. But note that the optimal strategy is to determine whether the tuple $\bm{x}=(x_1, x_2)$ came from the joint distribution $ab$ of independent $a$ and $b$ (in that order), or from $ba$ (opposite order). To do this we compute, given $\bm{x}$, the log likelihood ratio $l_{\frac{ab}{ba}}$ of $ab$ vs $ba$, which turns out to be simply related to the log likelihood ratios $l_{\frac{a}{b}}$ for the individual stimuli in the single-interval task:
\begin{gather*}
    \frac{p(ab | \bm{x})}{p(ba | \bm{x})} = \frac{p(\bm{x}|ab)}{p(\bm{x}|ba)} =\frac{p(x_1|a).p(x_2|b)}{p(x_1|b).p(x_2|a)} =\frac{p(a|x_1).p(b|x_2)}{p(b|x_1).p(a|x_2)}\\
    \implies l_{\frac{ab}{ba}} \left(x_1,x_2\right) = l_{\frac{a}{b}}(x_1)-l_{\frac{a}{b}}(x_2).
\end{gather*}

The optimal rule is to pick $ab$ if $l_{\frac{ab}{ba}} \left(x_1,x_2\right) >0$, i.e. if $l_{\frac{a}{b}}(x_1) > l_{\frac{a}{b}}(x_2)$. This is the familiar decision rule \cite{green1966signal}: the observer gets a log likelihood ratio from each distribution for the single-interval task (e.g. fig. \ref{fig:3_binary_class}b), and picks the larger likelihood ratio (not the larger stimulus).

When $a$ and $b$ are normals, $ab$ is the 2d normal with mean $\bm{\mu}_{ab}=(\mu_a, \mu_b)$ and sd $\mathbf{S}_{ab}=\text{diag} (\sigma_a, \sigma_b)$, and $ba$ is its flipped version (fig. \ref{fig:3_binary_class}c). The optimal decision rule (eq. \ref{eq:bdry_quadratic_coeffs}) boils down to selecting $ab$ when:
\begin{equation*}
    \left(\sigma_a^2-\sigma_b^2\right)\left(x_1^2-x_2^2\right)+2\left(\mu_a \sigma_b^2 - \mu_b \sigma_a^2\right)\left(x_1-x_2\right)>0.
\end{equation*}

When $\sigma_a = \sigma_b$, this is the usual condition of whether $x_1>x_2$. But when $\sigma_a \neq \sigma_b$, this optimal decision boundary comprises two perpendicular lines, solid and dashed (fig. \ref{fig:3_binary_class}c). The $x_1>x_2$ criterion is to use only the solid boundary, which is sub-optimal.

The minimum error rate $p(e)$ is the probability that the difference distribution of the two categories of fig. \ref{fig:3_binary_class}b exceeds 0. $l_{\frac{ab}{ba}}$ is the difference of scaled and shifted noncentral chi-squares $l_{\frac{a}{b}}$, so has generalized chi-square distributions for each category (fig. \ref{fig:3_binary_class}d), and we can calculate that $p(e)=p\left(\tilde{\chi}_{\bm{w}, \bm{k}, \bm{\lambda},0,0}\right)<0$, where
\begin{equation*}
    \bm{w}=\begin{bmatrix} \sigma_a^2 & -\sigma_b^2 \end{bmatrix},
    \; \bm{k}=\begin{bmatrix} 1 & 1 \end{bmatrix}, \; \bm{\lambda}=\frac{\mu_a-\mu_b}{\sigma_a^2-\sigma_b^2} \begin{bmatrix} \sigma_a^2 & \sigma_b^2 \end{bmatrix}.
\end{equation*}

If the two stimuli themselves arise from $k$-dimensional normals $N(\bm{\mu}_a,\mathbf{\Sigma}_a)$ and $N(\bm{\mu}_b,\mathbf{\Sigma}_b)$, then the optimal discrimination is between $2k$-dimensional normals $ab$ and $ba$, whose means are the concatenations of $\bm{\mu}_a$ and $\bm{\mu}_b$, and covariances are the block-diagonal concatenations of $\mathbf{\Sigma}_a$ and $\mathbf{\Sigma}_b$, in opposite order to each other.

\subsubsection{$m$-interval task}
Consider the $m$-interval ($m$-alternative forced choice) task with $m$ stimuli, one from the signal distribution $N(\mu_a,\sigma_a)$, and the rest from $N(\mu_b,\sigma_b)$. Following previous reasoning, the probability of the $i$th stimulus being the signal is an $m$-d normal, with mean vector whose $i$th entry is $\mu_a$ and the rest are $\mu_b$, and diagonal sd matrix whose $i$th entry is $\sigma_a$ and the rest are $\sigma_b$. The part of this log likelihood that varies across $i$ is:
\begin{gather*}
    -\sum_{j \neq i} \left(\frac{x_j-\mu_b}{\sigma_b}\right)^2 - \left(\frac{x_i-\mu_a}{\sigma_a}\right)^2 \\
    = -\underbrace{\sum_j \left(\frac{x_j-\mu_b}{\sigma_b}\right)^2}_\text{constant} + \underbrace{ \left(\frac{x_i-\mu_b}{\sigma_b}\right)^2 -\left(\frac{x_i-\mu_a}{\sigma_a}\right)^2 }_\text{varies with $i$}.
\end{gather*}
The optimal response is to pick the $m$-d normal with highest likelihood, i.e. pick the $x_i$ with the largest value of the second term above, i.e. with the largest log likelihood ratio $l$ of $a$ vs $b$, which is the familiar rule \cite{green1966signal}.

Analogous to Wickens \cite{wickens2002elementary} eq. 6.19, the maximum accuracy is then given by:
 \begin{equation*}
     p(c)= \int_{-\infty}^\infty F_b^{m-1} (l) \, f_a(l) \, dl
 \end{equation*}

where $f_a$ and $F_b$ are the pdf and cdf of $l$ under $a$ and $b$, which are known, so this can be evaluated numerically. For example, for 2, 3 and 4 intervals with the parameters of fig. \ref{fig:3_binary_class}a, the accuracy is 0.74 (fig. \ref{fig:3_binary_class}c), 0.64 and 0.58 (see example in the getting started guide for the toolbox). When the variances are equal, these computed accuracies  match table 6.1 of Wickens. \cite{wickens2002elementary}

\subsubsection{Discriminability index $d'$}
\label{sec:d_prime}
Bayesian classifiers are often used to model behavioral or neural performance in binary classification.  Within the Bayesian modeling framework, it is possible to estimate, from the pattern of errors, the separation (or overlap) of the decision variable distributions for the two categories, independent of the decision criterion (which may differ from the optimal value of zero). The discriminability index $d'$ measures this separation. If the two underlying distributions are equal-variance univariate normals $a$ and $b$, then $d'=\lvert \mu_a -\mu_b \rvert/\sigma$, and if they are multivariate with equal covariance matrices, then it is their Mahalanobis distance: $d'=\sqrt{(\bm{\mu}_a-\bm{\mu}_b)'\bm{\Sigma}^{-1}(\bm{\mu}_a-\bm{\mu}_b)} = \lVert \mathbf{S}^{-1}\bm{d} \rVert$, where $\bm{d}=\bm{\mu}_a-\bm{\mu}_b$ is the mean-difference vector.

We can show that this $d'$ in the full multi-dimensional space equals the one-dimensional $d'$ along the \textit{slice} of the space through the line $\bm{d}$ between the means. (This is different from the \textit{projection} of the space onto $\bm{d}$.) For this, we must compute the sd of the distributions on this slice $\bm{d}$. The radius of the error ellipsoid along any direction defines the sd of the slice of the distribution along that direction. From $\mathbf{S}$, we can calculate this sd in the direction of any unit vector $\hat{\bm{\eta}}$. Picture a vector from the ellipsoid center along $\hat{\bm{\eta}}$ to the edge of the ellipsoid, of length equal to the directional sd $\sigma_{\hat{\bm{\eta}}}$. In the normalized $\bm{z}$-space of unit spherical normals, this vector $\sigma_{\hat{\bm{\eta}}} \hat{\bm{\eta}}$ maps to a unit vector $\hat{\bm{u}}$ that is a radius of the unit spherical normal in some direction. So we can write:
\begin{equation*}
    \hat{\bm{u}} = \sigma_{\hat{\bm{\eta}}} \  \mathbf{S}^{-1}\hat{\bm{\eta}} \implies \lVert \hat{\bm{u}} \rVert = \sigma_{\hat{\bm{\eta}}} \lVert \mathbf{S}^{-1}\hat{\bm{\eta}} \rVert
    \implies \sigma_{\hat{\bm{\eta}}} = \frac{1}{\lVert \mathbf{S}^{-1}\hat{\bm{\eta}} \rVert}. 
\end{equation*}

(This is the sd of the \textit{conditional} distribution on this line, different from the sd of the \textit{marginal} distribution projected onto this line, which equals $\sqrt{\hat{\bm{\eta}}' \mathbf{\Sigma} \hat{\bm{\eta}}} = \lVert \mathbf{S} \hat{\bm{\eta}} \rVert$.) So we can say, along the unit vector $\hat{\bm{d}}$ in the direction $\bm{d}$:
\begin{equation*}
    \sigma_{\hat{\bm{d}}} = \frac{1}{\lVert \mathbf{S}^{-1}\hat{\bm{d}} \rVert} \implies d' = \lVert \mathbf{S}^{-1} \bm{d} \rVert = \frac{\lVert \bm{d} \rVert}{\sigma_{\hat{\bm{d}}}}, 
\end{equation*}

which is the mean-separation divided by the sd, i.e. the discriminability index, along the slice $\bm{d}$.

For bivariate distributions, we can calculate using the Mahalanobis distance that:
\begin{equation} \label{eq:corr_dprime}
    d'^2 =\frac{1}{1-\rho^2} \left(d'^2_x+d'^2_y-2\rho d'_x d'_y \right),
\end{equation}

where $\rho$ is the correlation coefficient, and here $d'_x=\frac{{\mu_b}_x-{\mu_a}_x}{\sigma_x}$ and $d'_y=\frac{{\mu_b}_y-{\mu_a}_y}{\sigma_y}$, i.e. including the signs of the mean differences instead of the absolute. In other words, the correlation reduces or increases the overall discriminability of the distributions, depending on their relative position. We will explain this visually in sec. \ref{sec:dprime_contribs}.

For univariate distributions with unequal variances, there exist several contending discriminability indices \cite{wickens2002elementary,chaddha1968empirical,simpson1973best}. A common one is Simpson and Fitter's $d'_a=\lvert \mu_a -\mu_b \rvert/\sigma_\text{rms}$ \cite{simpson1973best}, extended to general dimensions as the Mahalanobis distance using the pooled covariance, i.e. with $\mathbf{S}_\text{rms}=\left[\left(\bm{\Sigma}_a+\bm{\Sigma}_b\right)/2 \right]^\frac{1}{2}$ as the common sd. \cite{paranjpe1994selecting} Another index is Egan and Clarke's $d'_e=\lvert \mu_a -\mu_b \rvert/\sigma_\text{avg}$ \cite{egan1962psychophysics}, which we here extend to general dimensions using $\mathbf{S}_\text{avg}=\left(\mathbf{S}_a+\mathbf{S}_b\right)/2$.

\begin{figure}[!hb]
    \includegraphics[width=\columnwidth]{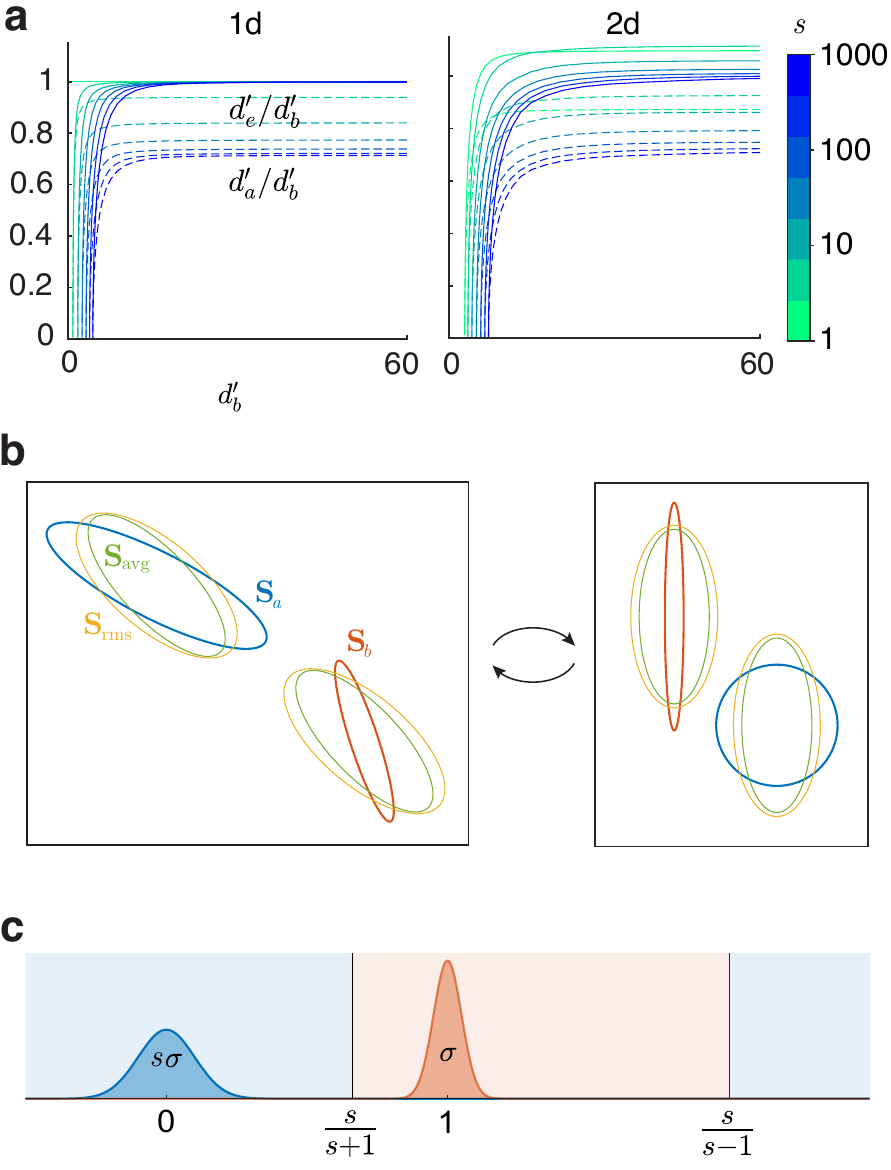}
    \caption{Comparing discriminability indices. \textbf{a.} Plots of existing indices $d'_a$ and $d'_e$ as fractions of the Bayes index $d'_b$, with increasing separation between two 1d and two 2d normals, for different ratios $s$ of their sd's. \textbf{b.} Left : two normals with 1 sd error ellipses corresponding to their sd matrices $\mathbf{S}_a$ and $\mathbf{S}_b$, and their average and rms sd matrices. Right: the space has been linearly transformed, so that $a$ is now standard normal, and $b$ is aligned with the coordinate axes. \textbf{c.} Discriminating two highly-separated 1d normals.}
    \label{fig:4_discriminability}
\end{figure}

These unequal-covariance measures are simple approximations that do not describe the exact separation between the distributions. However, our methods can be used to define a discriminability index that exactly describes the separation between two arbitrary distributions (even non-normal). First, we determine the minimum possible (Bayes) error when $\mathbf{V}=\bm{1}$ and priors are equal. This is $p(e) = (p(B | a)+p(A | b))/2$, where we can also write these error rates in terms of the distributions $f_a$ and $f_b$ of the log likelihood ratios:
\begin{align*}
    p(B | a) &= \int_{-\infty}^0 f_a(l) \ dl = F_a(0), \\
    p(A | b) &= \int_0^{\infty} f_b(l) \ dl = 1- F_b(0).
\end{align*}
(For 1d normals, these are given by eq. \ref{eq:1d_class}.) 

The Bayes error rate is also the overlap area of the distributions. If $p_a(\bm{x})$ and $p_b(\bm{x})$ are the two distributions weighted by their priors, so that their masses together sum to 1, then the Bayes error rate is their overlap area (e.g. the overlap area in fig. \ref{fig:3_binary_class}a), which is the area under the function which is the minimum of the two distributions:
\begin{equation*}
    p(e) = \int \min(p_a(\bm{x}),p_b(\bm{x})) \ d\bm{x}.
\end{equation*}

If we know the optimal boundary between two distributions, we can compute $p(e)$ by computing the masses on either side of it that add to the error rate. But if we do not know the optimal boundary, e.g. when we have empirically sampled histograms, we can compute $p(e)$ using the area of overlap above, but the value will vary with our choice of histogram bins.

We now define the \textit{Bayes discriminability index} as the equal-variance index that corresponds to this same Bayes error, i.e. the separation between two unit variance normals that have the same overlap as the two distributions, which comes out to be twice the $z$-score of the maximum accuracy:
\begin{align*}
    d'_b &=-2Z\left(\text{Bayes error / overlap fraction } p\left(e\right)\right) \\
    &=2Z\left(\text{best accuracy }p\left(c\right)\right)
\end{align*}

This index is the best possible discriminability, i.e. by an ideal observer. It extends to all cases as a smooth function of the layout and shapes of the distributions, and reduces to $d'$ for equal variance/covariance normals.

$d'_b$ is a positive-definite statistical distance measure that is free of assumptions about the distributions, like the Kullback-Leibler divergence $D_\text{KL}(a,b)$, which is the expected log likelihood ratio $l$ under the $a$ distribution (mean of the blue distributions in figs. \ref{fig:3_binary_class}b and d). $D_\text{KL}(a,b)$ is asymmetric, whereas $d'_b(a,b)$ is symmetric for the two distributions. However, $d'_b$ does not satisfy the triangle inequality. For example, consider three equal-width, consecutively overlapping uniform distributions: $a$ over $[0,3]$, $b$ over $[2,5]$, and $c$ over $[4,7]$. $b$ overlaps with $a$ and $c$: $d'_b(a,b)=d'_b(b,c)=2Z(2/3)$, but $a$ and $c$ do not overlap: $d'_b(a,c)=\infty \nless d'_b(a,b)+d'_b(b,c)$.

In fig. \ref{fig:4_discriminability}a, we compare $d'_b$ with $d'_a$ and $d'_e$ for different mean-separations and sd ratios of two normals, in 1d and 2d. We first take two 1d normals and increase their discriminability by equally shrinking their sd's while maintaining their ratio $\sigma_a/\sigma_b=s$, i.e. effectively separating the means. We repeat this by starting with two 2d normals with different sd matrices, one of them scaled by different values $s$ each time, then shrink them equally.

Extending previous findings \cite{simpson1973best}, we see that in 1d (fig. \ref{fig:4_discriminability}a left), $d'_a \leq d'_e \leq d'_b$. Thus, $d'_a$ and $d'_e$ underestimate the optimal discriminability of normal distributions. The worst case is when the means are equal, so $d'_a=d'_e=0$, but $d'_b$ is positive, since unequal variances still provide discriminability.

Now consider the opposite end, where large mean-separation has a much greater effect on discriminability than sd ratios. Even here, the underestimate by $d'_a$ persists, and worsens as the sd's become more unequal, reaching nearly 30\% in the worst case. $d'_e$  is a better estimate throughout, and equals $d'_b$ at large separation.

In higher dimensions, $d'_a \leq d'_e$ still, and they still usually underestimate $d'_b$ (especially when means are close), but there are exceptions (fig. \ref{fig:4_discriminability}a right, and fig. \ref{fig:2_examples}d).

We can theoretically show that $d'_a \leq d'_e$ in all dimensions and cases. In 1d, this is simply because $\sigma_\text{avg} \leq \sigma_\text{rms}$, and at the limit of highly unequal sd's, $\sigma_\text{avg}/\sigma_\text{rms} \rightarrow 1/\sqrt{2}$, so $d'_a \rightarrow d'_e/\sqrt{2}$, which is the 30\% underestimate. In higher dimensions, we can show analogous results using fig. \ref{fig:4_discriminability}b as an example. The left figure shows two normals with error ellipses corresponding to their sd's, and their average and rms sd's. Now we make two linear transformations of the space: first we standardize normal $a$, then we diagonalize normal $b$ (i.e. a rotation that aligns the axes of error ellipse $b$ with the coordinate axes). In this space (right figure), $\mathbf{S}_a=\mathbf{1}$, $\mathbf{S}_b$ is diagonal, and the axes of $\mathbf{S}_\text{avg}$ and $\mathbf{S}_\text{rms}$ are the average and rms of the corresponding axes of $\mathbf{S}_a$ and $\mathbf{S}_b$. $\mathbf{S}_\text{rms}$ is hence bigger than $\mathbf{S}_\text{avg}$, so has larger overlap at the same separation, so $d'_a \leq d'_e$. The ratio of $d'_a$ and $d'_e$ is $\lVert \mathbf{S}_\text{rms}^{-1}\bm{d} \rVert/\lVert \mathbf{S}_\text{avg}^{-1}\bm{d} \rVert$, the ratio of the 1d slices of the average and rms sd's along the axis through the means. When these are highly unequal, we again have $d'_a \rightarrow d'_e/\sqrt{2}$ in general dimensions.

We can also show that at large separation in 1d, $d'_e$ converges to $d'_b$. Consider normals at 0 and 1 with sd's $s \sigma$ and $\sigma$ (fig. \ref{fig:4_discriminability}c). At large separation ($\sigma \rightarrow 0$), the boundary points, where the distributions cross, are $\frac{s}{s \pm 1}$. The right boundary is $\frac{1}{\sigma(s-1)}$ sd's from each normal, so it adds as much accuracy for the left normal as it subtracts for the right. So only the inner boundary is useful, which is $\frac{1}{\sigma(s+1)}$ sd's from each normal. The overlap here thus corresponds to $d'_b=\frac{2}{\sigma(s+1)}=d'_e$. So, when two 1d normals are too far apart to compute their overlap (see performance section) and hence $d'_b$, the toolbox returns $d'_e$ instead.

Given that $d'_e$ is often the better approximation to the best discriminability $d'_b$, why is $d'_a$ used so often? Simpson and Fitter \cite{simpson1973best} argued that $d'_a$ is the best index, because it is the accuracy in a two-interval task with stimuli $x_1$ and $x_2$, using the criterion $x_1>x_2$. But as we saw, this is not the optimal way to do this task. The optimal error $p(e)$ is instead as calculated previously, and $d'_b(ab,ba)=2Z\left(1-p\left(e\right)\right)$ is the best discriminability. Unfortunately, this does not have a simple relationship with $d'_b(a,b)$ for the yes/no task. But we can calculate mathematically here that $d'_e(ab,ba)= \sqrt{2}d'_e(a,b)$, which may still better approximate the best discriminability than $d'_a(ab,ba)=\sqrt{2}d'_a(a,b)$.

A brief note about Grey and Morgan's approximate index, which uses the geometric mean of the sd's: this behaves inconsistently; it underestimates $d'_b$ at small discriminability, but overestimates it at large discriminability.

In sum, $d'_b$ is the maximum discriminability between normals in all cases, including two-interval tasks, especially when means are closer and variances are unequal. $d'_e$ often approximates it better than $d'_a$, e.g. when the decision variable in a classification task is modeled as two unequal-variance 1d normals.

\subsubsection{ROC curves}

\begin{figure}[!t]
    \includegraphics[width=\columnwidth]{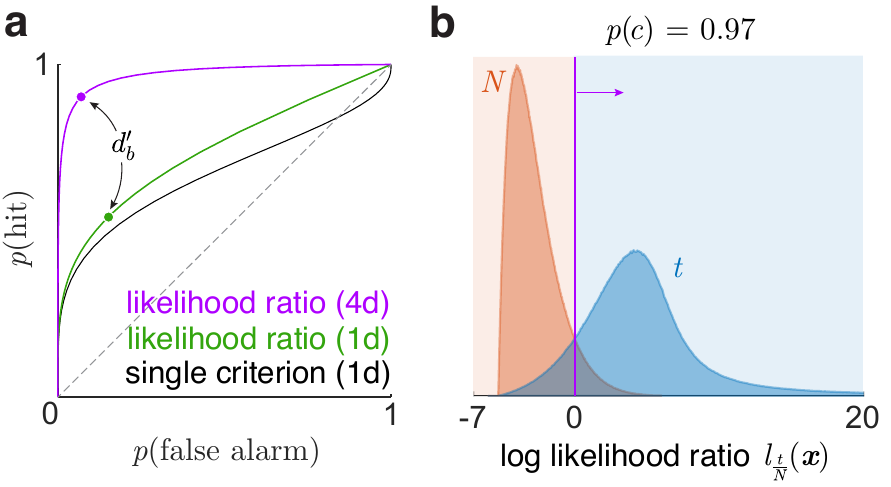}
    \caption{ROC curves. \textbf{a.} Yes/no ROC curves for a single shifting criterion (black), vs. a shifting likelihood-ratio (green), between the two 1d normals of fig. \ref{fig:3_binary_class}a (adapted from Wickens \cite{wickens2002elementary}, fig. 9.3), and a shifting likelihood-ratio between a normal and a $t$ distribution in 4d (purple). The optimal two-interval accuracies of the 1d normals (fig. \ref{fig:3_binary_class}c) and the 4d distributions (fig. \ref{fig:6_ROC}b) are 0.74 and 0.97, equal to the areas under their likelihood-ratio curves here. The points marked on these curves are the farthest from the diagonal, and correspond to the Bayes discriminability. \textbf{b.} Distributions of the log likelihood ratio of the 4d $t$ vs normal distribution. Sweeping the criterion corresponds to moving along the purple ROC curve of a.}
    \label{fig:6_ROC}
\end{figure}

ROC curves track the outcome rates from a single criterion swept across two 1d distributions (e.g. black curve of fig. \ref{fig:6_ROC}a), or varying the likelihood-ratio between any two distributions in any dimensions (green and purple curves), which corresponds to sweeping a single criterion across the 1d distributions of the likelihood ratio $l$ (figs. \ref{fig:3_binary_class}b and \ref{fig:6_ROC}b for the green and purple curves).

Discriminability indices are frequently estimated from ROC curves. $d'_a$ is $\sqrt{2}$ times the $z$-score of the single-criterion ROC curve area. $d'_b$ has no such simple relationship with curve area, but can be estimated in different ways. Even though $d'_b$ uses both criteria for unequal-variance 1d normals, it can still be estimated from the usual single-criterion ROC curve. Assume that the normals are $N(0,1)$ and $N(\mu,\sigma^2)$. From the single-criterion ROC curve, we first estimate $\mu$ and $\sigma$, then we use our method to compute $d'_b$ of normals with these parameters.

$d'_b$ can also be estimated from a likelihood-ratio ROC curve. For any two distributions in any dimensions, $d'_b$ corresponds to the accuracy at the point along their likelihood-ratio ROC curve that maximizes $p(\text{hit})-p(\text{false alarm})$, which is the farthest point from the diagonal, where the curve tangent is parallel to the diagonal (fig. \ref{fig:6_ROC}a).

\subsubsection{Custom classifiers}
Sometimes, instead of the optimal classifier, we need to test and compare sub-optimal classifiers, e.g. one that ignores a cue, or some cue covariances, or a simple linear classifier. So the toolbox allows the user to extract the optimal boundary and change it, and explicitly supply some custom sub-optimal classification boundary. Fig. \ref{fig:2_examples}d compares the classification of two bivariate normals using the optimal boundary (which corresponds to $d'_b$), vs. using a hand-supplied linear boundary. Just as with integration, one can supply these custom classification domains in quadratic, ray-trace or implicit form, and use set operations on them.

\subsection{Classifying using data}
If instead of normal parameters, we have labelled data as input, we can estimate the parameters. The maximum-likelihood estimates of means, covariances and priors of normals are simply the sample means, covariances and relative counts. With these parameters we can compute the optimal classifier $\beta(\bm{x})$ and the error matrix. We can further calculate another quadratic boundary $\gamma(\bm{x})$ to better separate the given samples: starting with $\beta(\bm{x})$, we optimize its $(k+1)(k+2)/2$ independent parameters to maximize the classification outcome value of the given samples. This is important for non-normal samples, where the optimal boundary between estimated normals may not be a good classifier. This optimization then improves classification while still staying within the smooth quadratic family and preventing overfitting. Fig. \ref{fig:2_examples}e shows classification based on labelled non-normal samples.

If, along with labelled samples, we supply a custom quadratic classifier, the toolbox instead optimizes this for the sample. This is useful, say, in the following case: suppose we have already computed the optimal classifier for samples in some feature space. Now if we augment the data with additional features, we may start from the existing classifier (with its coefficients augmented with zeros in the new dimensions) to find the optimal classifier in the larger feature space.

\subsection{Scaling the discriminability of data}

\begin{figure}[!b]
    \includegraphics[width=\columnwidth]{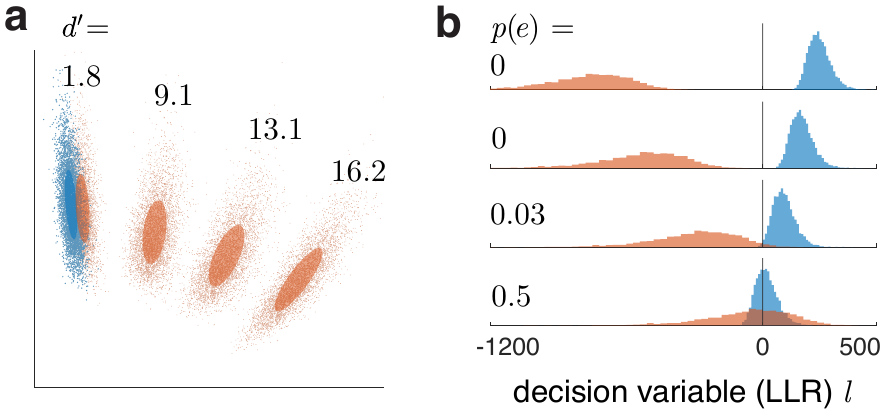}
    \caption{Scaling the discriminability of two distributions by linear interpolation. \textbf{a.} The discriminability of two general distributions (the orange one is non-normal) is scaled by linearly interpolating the mean and sd matrix of the orange towards the blue. Ellipses show their error ellipses. \textbf{b.} Scaling the discriminability of the data distributions in plot a by computing their decision variables under the normal model, and pulling them towards 0. $p(e)$ is the error rate.}
    \label{fig:7_dprime_scaling}
\end{figure}

Consider two samples $\bm{x}_a$ and $\bm{x}_b$, with means $\bm{\mu}_a$, $\bm{\mu}_b$ and covariances $\bm{\Sigma}_a$, $\bm{\Sigma}_b$ (sd matrices $\mathbf{S}_a$, $\mathbf{S}_b$). In some cases, we may want to scale their discriminability by moving them closer or farther apart. One such case is when we are modeling a detection or classification task, and the performance of the model or ideal observer exceeds that of the subject or observed data. In that case, we can move the model variable distributions closer together so that it matches the observed performance, while also predicting which \textit{specific data points} should start overlapping and be misclassified.

To do this, we first model the samples as multinormals: $\bm{x}_a \sim \mathbf{S}_a\bm{z}+\bm{\mu}_a$ and $\bm{x}_b \sim \mathbf{S}_b\bm{z}+\bm{\mu}_b$. Now we effect transformations to linearly interpolate these multinormals closer together. There are different flavours of doing this depending on the case. If we suppose that $a$ is the noise distribution, then it should stay fixed while the signal distribution $b$ is pulled towards it. So we introduce the interpolation factor $r$, such that the parameters of distribution $b$ become: $\bm{\mu}'_b=r \bm{\mu}_b+(1-r)\bm{\mu}_a$ and $\mathbf{S}'_b=r \mathbf{S}_b+(1-r)\mathbf{S}_a$. As $r$ goes to 0, distribution $b$ linearly morphs into $a$. We can effect this transformation on the data points $\bm{x}_b$ by first whitening them: $\bm{z}_b = \mathbf{S}_b^{-1}(\bm{x}_b-\bm{\mu}_b)$, then linearly transforming them to the new parameters: $\bm{x}'_b = \mathbf{S}'_b\bm{z}_b+\bm{\mu}'_b$. We can also have other flavours, such as interpolating the parameters of the two distributions towards their common means. One important thing to note here is that the distribution parameters that we should linearly interpolate are the mean vector and the \textit{sd matrix}, not the covariance matrix, because the points are a linear function of these parameters.

Fig. \ref{fig:7_dprime_scaling}a shows two distributions where this linear warping is implemented. These can be general distributions. Here, for example, the blue one is normal while the orange one isn't.  The orange distribution is interpolated towards the blue in steps from $r=1$ through 0.05, and the $d'$ shrinks. Using the optimal classification boundary at any step in the scaling will tell us which specific data points would be misclassified in the model, which we can then compare with our subject or observed data.

A different way that we can scale the discriminability of two data distributions in one dimension instead of many, is by computing their decision variables (log likelihood ratio $l$ that a point belongs to $a$ vs $b$) under the normal model, then moving them closer together or farther apart. Fig. \ref{fig:7_dprime_scaling}b, top subplot, shows the two distributions of the Bayes decision variable for the two original bivariate samples in plot a that were farthest apart. (The blue sample $a$ is now on the right because its log likelihood ratio is positive.) If the median value of a decision variable distribution is $l_m$, we can map the decision variables to $l'=l-l_m(1-r)$. As the scale factor $r$ goes from 1 to 0, each decision variable distribution is pulled in until half of its points are on either side of 0, i.e. the error rate goes to chance and the discriminability goes to 0. Once again, we can use this to scale down the discriminability of a model to match a subject or observed data, and see if the model predicts the specific trials or data points that were misclassified.

\subsection{Multiple normals}
The optimal classifier between two normals is a quadratic, so error rates can be computed using the generalized chi-square method or the ray-trace method. When classifying amongst more than two normals, the decision region for each normal is the intersection of its quadratic decision regions $q^i_n(\bm{x})>0$ with all the other normals $i$, and may be written as:
\begin{equation*}
    f(\bm{x})=\min_{i} q^i_n(\bm{x}) > 0.
\end{equation*}
This is not a quadratic, so only the ray-trace method can compute the error rates here, by using the intersection operation on the domains as described before. Fig. \ref{fig:2_examples}f shows the classification of several normals with arbitrary means and covariances.

\subsection{Combining and reducing dimensions}
It is often useful to combine the multiple dimensions in a problem to fewer, or one dimension \cite{orucc2003weighted}. Mapping many-dimensional integration and classification problems to fewer dimensions allows visualization, which can help us understand multivariate normal models and their predictions, and to check how adequately they represent the empirical or other theoretical probability distributions for a problem.

As we have described, the multi-dimensional problem of integrating a normal probability in the domain $f(\bm{x})>0$ can be viewed as the 1d integral of the pdf of $f(\bm{x})$ above 0. Similarly, multi-dimensional binary classification problems with a classifier $f(\bm{x})$ can be mapped to a 1d classification between two distributions of the scalar decision variable $f(\bm{x})$, with the criterion at 0, while preserving all classification errors. For optimally classifying between two normals, mapping to the Bayes decision variable $\beta(\bm{x})$ is the optimal quadratic combination of the dimensions. For integration and binary classification problems in any dimensions, the toolbox can plot these 1d 'function probability' views (fig. \ref{fig:2_examples}g). With multiple classes, there is no single decision variable to map the space to, but the toolbox can plot the projection along any chosen vector direction. Fig. \ref{fig:2_examples}h shows the classification of samples from four 4d $t$ distributions using normal fits, projected onto the axis along (1,1,1,1).

For a many-dimensional classification problem, we can also define a decision variable on a subset of dimensions to combine them into one, then combine those combinations further etc, according to the logic of the problem.

In the sections below, we shall see examples of such applications, where we map to fewer dimensions to see how well a multivariate normal model works for a problem, and also combine groups of cues to organize a problem and get visual insight.

\subsection{Testing a normal model for classification}

\begin{figure}[!b]
    \includegraphics[width=\columnwidth]{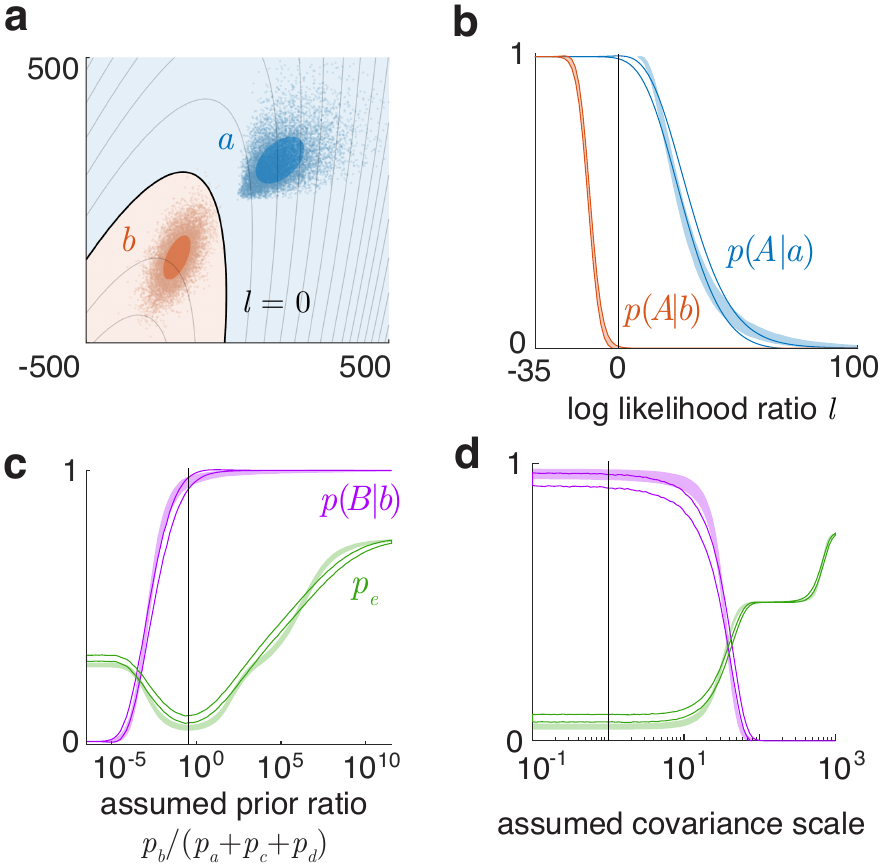}
    \caption{Testing normal approximations for classification. \textbf{a.} Classifying two empirical distributions ($a$ is not normal). Gray curves are contours of $l$, i.e. the family of boundaries corresponding to varying likelihood ratios of the two fitted normals. \textbf{b.} Mean $\pm$ sd of hit and false alarm fractions observed (color fills) vs. predicted by the normal model (outlines), along this family of boundaries. Vertical line is the optimal boundary. \textbf{c.} Similar bands for class $b$ hits and overall error, for the 4d 4-class problem of fig. \ref{fig:2_examples}h, across boundaries assuming different priors $p_b$, and \textbf{d.} across boundaries assuming different covariance scales ($d'$s).}
    \label{fig:8_normality_check}
\end{figure}

The results developed here are for normal distributions. But even when the variables in a classification problem are not exactly normal (e.g. either they are an empirical sample, or they are from some known, but non-normal distribution), we can still use the current methods if we check whether normals are an adequate model for them. One test, as described before, is to project the distributions to one dimension, either by mapping to a quadratic form (fig. \ref{fig:2_examples}g), or to an axis (fig. \ref{fig:2_examples}h), where we can visually compare the projections of the observed distributions and those of their fitted normals. 

We could further explicitly test the normality of the variables with measures like negentropy, but this is stricter than necessary. If the final test of the normal model is against outcomes of a limited-trials classification experiment, then it is enough to check for agreement between outcome counts predicted by the true distributions and their normal approximations, given the number of trials. For any classification boundary, we can calculate outcome rates, e.g. $p(A|a)$ for a hit, determined from the true distributions, vs. from the normal approximations. The count of hits in a task is binomial with parameters equal to the number of $a$ trials and $p(A|a)$, so we can compare its count distribution between the true and the normal model.

If the classes are well-separated (e.g. for ideal observers), the optimal boundary provides near-perfect accuracy on both the true and the normal distributions, so comparing yields no insight. To make the test more informative, we repeat it as we sweep the boundary across the space into regions of error, to show if the normal model still stands. This is similar to how the decision criterion between two 1d distributions is swept to create an ROC curve that characterizes the classification more richly than a single boundary. In multiple dimensions, there is more than one unique way to sweep the boundary. We pick two common suboptimal boundary families. The first corresponds to an observer being biased towards one type of error or another, i.e. a change in the assumed ratio of priors or outcome values. The second is an observer having an internal discriminability different from the true (external) one (e.g. due to blurring the distributions by internal noise), so adopting a boundary corresponding to covariance matrices that are scaled by a factor. When there are two classes, the boundaries for both of these sub-optimal observers correspond to a shift in the constant offset $q_0$ (eq. \ref{eq:bdry_quadratic_coeffs}), i.e. a shift in the likelihood ratio of the two normals. So we are simply moving along the normal likelihood-ratio ROC curves, as we compare the outcome rates of the true and the normal distributions.

Fig. \ref{fig:8_normality_check}a shows the classification of two empirical distributions, where $a$ is not normal, and gray curves show this family of boundaries, which are simply contours of the log likelihood ratio $l$. Since $a$ and $b$ are well-separated, the ROC curves for both true and normal distributions would hug the top and left margins, so they cannot be compared. Instead, we detach the hits and false alarms from each other and plot them individually against the changing likelihood ratio criterion, which gives us more insight. Fig. \ref{fig:8_normality_check}b shows the mean $\pm$ sd bands of hits and false alarms from applying these boundaries on samples of 100 trials (typical of a psychophysics experiment) from each true distribution, vs. the normal approximations. They exactly coincide for false alarms / correct rejections, but deviate for hits / misses, correctly reflecting that $b$ is normal but $a$ is not. The investigator can judge if this deviation is small enough to be ignored for their problem.

Now consider the case of applying these tests to multi-class problems. The two kinds of sub-optimal boundaries we picked are no longer the same family here. Recall that the classification problem of \ref{fig:2_examples}h had four 4d $t$ distributions. Fig. \ref{fig:8_normality_check}c shows similar tests to see if this problem (with priors now equal) are well-modeled by normals. The family of boundaries corresponds to varying the assumed prior $p_b$. We may compare any of the 16 outcome rates here, e.g. $p(B|b)$, and also the overall error $p(e)$. When there are multiple classes, for any given true class, the numbers of responses in the different classes are multinomially distributed, so that the total number of wrong responses is again binomially distributed. $p(e)$ is the prior-weighted sum of these binomially distributed individual errors, so we can calculate its mean and sd predicted by the observed vs. the normal distributions. Fig. \ref{fig:8_normality_check}d shows the test across boundaries corresponding to all covariance matrices scaled by a factor, changing the $d'$ between the classes. 

Some other notable suboptimal boundaries to consider for this test are ones that correspond to adding independent noise to the cues (which changes only their variances but not their covariances), ones that ignore certain cues or cue covariances, or simple flat boundaries. As seen here, even for many-dimensional distributions that cannot be visualized, these tests can be performed to reveal some of their structure, and to show which specific outcomes deviate from normal prediction for which boundaries. 

When the problem variables have a known non-normal theoretical distribution, the maximum-likelihood normal model is the one that matches its mean and covariance, and these tests can be performed by theoretically calculating or bootstrap sampling the error rate distributions induced by the known true distributions.

\section{Matlab toolbox: functions and examples}

For an integration problem, the toolbox provides the function \code{integrate\_normal} that inputs the normal parameters  and the integration domain (as quadratic domain coefficients, limits of a rectangular domain, a ray-trace domain function or an implicit domain function), and outputs the integral and its complement, the boundary points computed, and a plot of the normal probability or function probability view. The toolbox also provides functions \code{norm\_fun\_pdf}, \code{norm\_fun\_cdf} and \code{norm\_fun\_inv} to compute pdf's, cdf's and inverse cdf's of functions of (multi)normals. The function \code{classify\_normals} for a binary classification problem inputs normal parameters, priors, outcome values and an optional classification boundary, and outputs the coefficients of the quadratic boundary $\beta$ and points on it, the error matrix and discriminability indices $d'_b$, $d'_a$ and $d'_e$, and produces a normal probability or function probability plot. Optionally, it can also return the contributions to $d'$ from each dimension along with a divided color-bar in the plot. With sample input, it first estimates the normal parameters (means, covariances and priors) and returns the optimal boundary $\beta$, and also which sample points were correctly classified. In addition, it returns the coefficients of the sample-optimal boundary $\gamma$ and points on it, error matrices and $d'_b$ values of classifying the samples using $\beta$ and $\gamma$, and the mapped scalar decision variables $\beta(\bm{x})$ and $\gamma(\bm{x})$ from the samples. The ability of scaling the $d'$, by either warping the multivariate distributions, or their scalar decision variables, is available in this function in the options \code{d\_scale} and \code{d\_scale\_type}. The function \code{classify\_normals\_multi} for classifying multiple normals.

Many different example problems, including every problem discussed in this paper (examples in figs. \ref{fig:2_examples}, \ref{fig:3_binary_class}, \ref{fig:4_discriminability}, \ref{fig:6_ROC} and \ref{fig:7_dprime_scaling}, tests in figs. \ref{fig:8_normality_check} and \ref{fig:9_performance}, and research applications in fig. \ref{fig:10_examples_vision}) are available as interactive demos in the `getting started' live script of the toolbox, and can be easily adapted to other problems.

\section{Performance benchmarks}

\begin{figure}[!t]
    \includegraphics[width=\columnwidth]{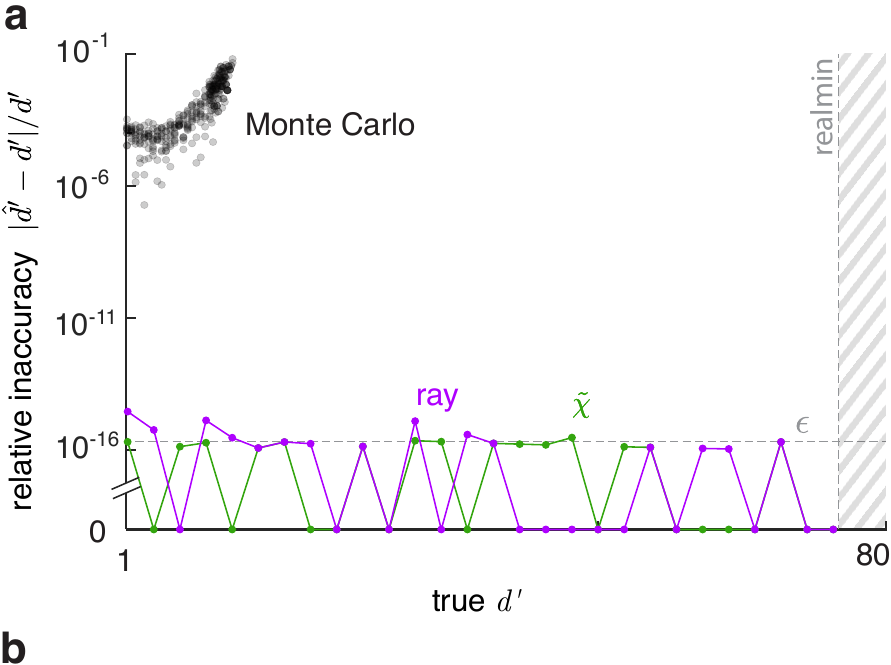}

  \begin{center} 
  \begin{tabular}{|l|c|c|c|c|c|}
    \hline
    \multirow{2}{0cm}{Problem} & \multicolumn{2}{c|}{Rel. diff. from MC} & \multicolumn{3}{c|}{Time (s)} \\
    \cline{2-6}
    & $\tilde{\chi}$ & ray & $\tilde{\chi}$ & ray & MC\\
    \hline
    \ref{fig:2_examples}a top, $p$ & n/a & 0 & n/a & 0.9 & 0.02 \\ 
    \ref{fig:2_examples}a bottom, $p$ & n/a & 0 & n/a & 0.3 & 0.003 \\ 
    \ref{fig:2_examples}d, $p(e)$ & 0 & 0 & 0.7 & 0.1 & 0.05 \\ 
    \ref{fig:2_examples}d, $p'(e)$ & 0 & 0 & 0.006 & 0.08 & 0.04  \\ 
    \ref{fig:2_examples}e, 1\textsuperscript{st} $p(e)$ & 0 & 0 & 1.7 & 0.12 & 0.01 \\ 
    \ref{fig:2_examples}g, $p$ & 0 & 8e-4 & 0.4 & 1.7 & 0.02  \\ 
    \ref{fig:2_examples}g, 1\textsuperscript{st} $p(e)$ & 0 & 9e-4 & 0.7 & 1.4 & 0.01 \\ \hline
  \end{tabular}
    \end{center}

  \caption{Performance benchmarks of the generalized chi-square (denoted by $\tilde{\chi}$) and ray-trace methods, against a standard Monte-Carlo method. \textbf{a.} Relative inaccuracies in $d'$ estimates by the Monte-Carlo method (across multiple runs), and our two methods, as the true $d'$ increases. Monte-Carlo estimates that take similar time rapidly become erroneous, failing beyond $d'\approx 10$. Our methods stay extremely accurate (around machine epsilon $\epsilon$) up to $d' \approx 75$, which corresponds to the smallest error rate representable in double precision (`realmin'). \textbf{b.} For several problems of fig. \ref{fig:2_examples}, relative differences in the outputs of the two methods from the Monte Carlo estimate, and computation times for 1\% precision.}
  \label{fig:9_performance}
\end{figure}

In this section we test the performance (accuracy and speed) of our Matlab implementations of the generalized chi-square and ray-trace algorithms, against a standard Monte Carlo integration algorithm. We first set up a case where we know the ground truth, and compare the estimates of all three methods at the limit of high discriminability (low error rates) where it is most challenging (which occurs for computational models and ideal observers). We take two 3d normals with the same covariance matrix, so that the true discriminability $d'$ is exactly calculated as their Mahalanobis distance. Now we increase their separation, while computing the optimal error with each of our methods at maximum precision, and a discriminability from it. The generalized chi-square method is very fast (due to the trivial planar boundary here), and the ray-trace method takes an average of 40s. For a fair comparison, we use $10^8$ samples for the Monte Carlo, which also takes $\sim$40s. Each method returns an estimate $\hat{d}'$. Fig. \ref{fig:9_performance}a shows the relative inaccuracies $\lvert \hat{d}' -d' \rvert / d'$ as true $d'$ increases.
With increasing separation, the Monte Carlo method quickly becomes inaccurate, since the error rate, i.e. the probability content in the integration domain, becomes too small to be sampled. The method stops working beyond $d' \approx 10$, where none of the $10^8$ samples fall in the error domain. In contrast, inaccuracies in our methods are extremely small, of the order of the double-precision machine epsilon $\epsilon$, demonstrating that the algorithms contain no systematic error besides machine imprecision (however, Matlab's native integration methods may not always reach the desired precision for a problem). This is possible because a variety of techniques are built into our algorithms to preserve accuracy, such as holding tiny and large summands separate to prevent rounding, using symbolic instead of numerical calculations, and using accurate tail probabilities. The inaccuracies do not grow with increasing separation, until $d' \approx 75$, which corresponds to the smallest error rate $p(e)$ representable in double-precision (\code{realmin} = 2e-308), beyond which both methods return $p(e)=0$ and $d'_b=\infty$. For 1d problems beyond this, we can use $d'_e$ instead. In a later paper \cite{das2024new}, we use logarithmic approximations or variable precision arithmetic in the ray method to extend the computation of $d'$ even beyond 75.

Next, we compare the three methods across several problems of fig. \ref{fig:2_examples}. The values here are large enough that Monte Carlo estimates are reliable and quick, so we use it as a provisional ground truth. We compute the values with all three methods up to maximum practicable precisions, then we calculate the relative (fractional) differences of our methods from the Monte Carlo. If a value is within the spread of the Monte Carlo estimates, we call the relative difference 0. Table \ref{fig:9_performance}b lists these, along with the times to compute the values to 1\% precision on an AMD Ryzen Threadripper 2950X 16-core processor. We see that both of our methods produce accurate values at comparable speeds.

\section{Applications in visual detection}

\begin{figure}
    \includegraphics[width=\columnwidth]{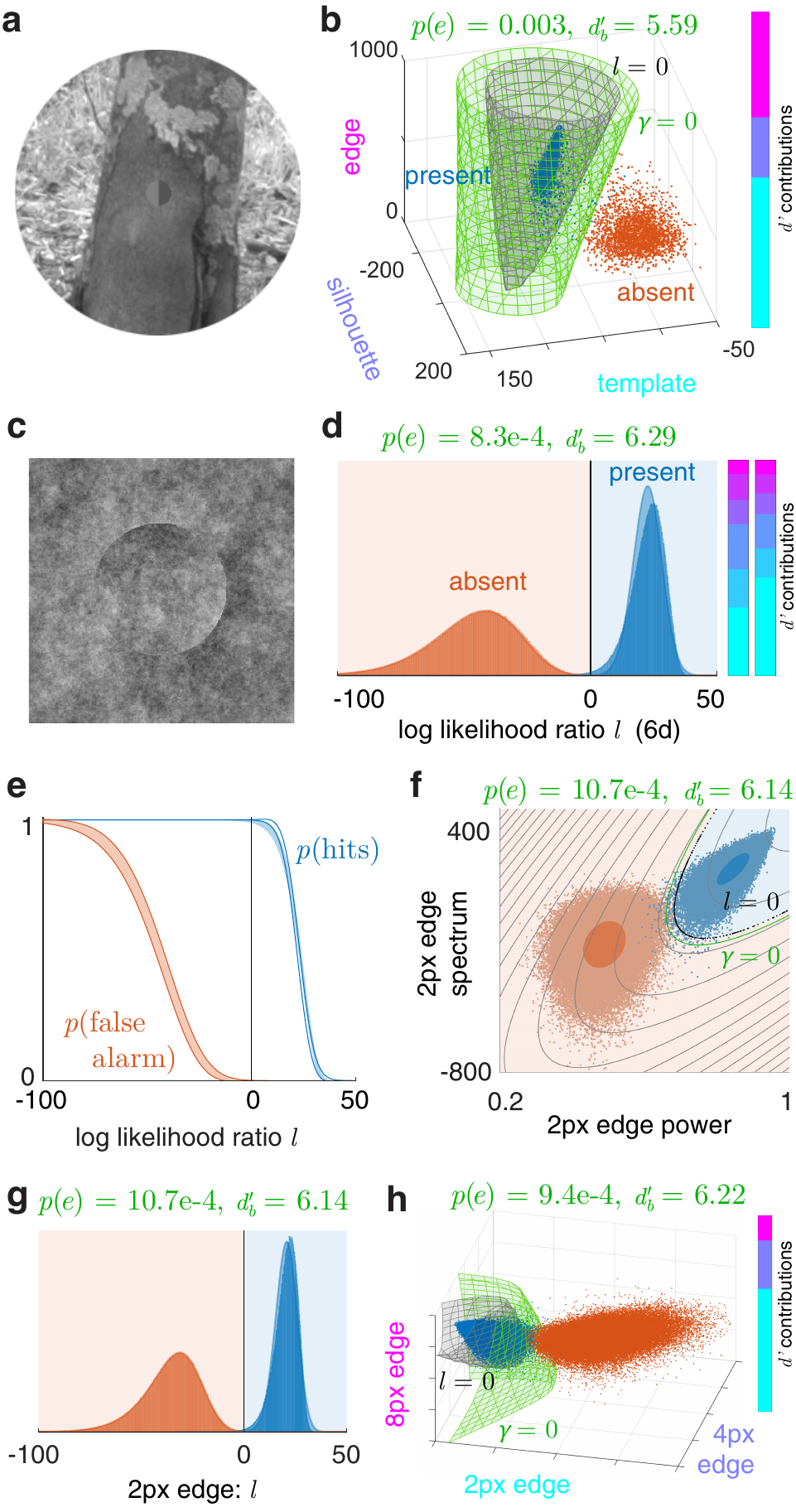}
    \caption{Applying the method and toolbox to visual target detection studies. \textbf{a.} Example target on a natural background. \textbf{b.} Classifying such images with the target present or absent, using 3 features. $p(e)$ and $d'_b$ correspond to the classification error of the sample points using $\gamma$. Color-bar shows the relative contributions $\delta^{\text{rem}}_i$ of the features to the overall $d'$, which account for their correlations (\S \ref{sec:dprime_contribs}). \textbf{c.} Example camouflage detection image. \textbf{d.} Classifying these using 6 features, viewed in terms of the log likelihood ratio $l$. Left color-bar shows the proportions of the individual $d'_i$ of each feature, right one shows their relative contributions $\delta^{\text{rem}}_i$. The first feature, edge power at 2px, is the most helpful, particularly when we consider correlations. \textbf{e.} Bootstrap mean $\pm$ sd of hit and false alarm fractions from applying a family of boundaries (corresponding to varying the criterion likelihood ratio) on 100 samples of the true 6d cue distributions (color fills), vs. their normal approximations (outlines). \textbf{f.} Classifying with only two cues computed at 2px. Grey curves are contours of the log likelihood ratio $l$. \textbf{g.} Combining the two cues of plot f into one using $l$ (i.e. the space of plot f has been projected along the gray contours). \textbf{h.} Classifying with such combined cues at 3 scales. Color-bar of $\delta^{\text{rem}}_i$ contributions shows that the 2px edge aids the most in detection.}
    \label{fig:10_examples_vision}
\end{figure}

We demonstrate the use of these methods in visual detection tasks that have multiple cues with different variances and correlations.

\subsection{Detecting targets in natural scenes}

We have applied this method in a study to measure how humans compare against a nearly ideal observer in detecting occluding targets against natural scene backgrounds in a variety of conditions \cite{walshe2020detection}. We placed a target on a random subset of natural images, then blurred and downsampled them to mimic the effect of the early visual system (fig. \ref{fig:10_examples_vision}a). We sought to measure how well the targets on these degraded images can be detected using three cues: related to the luminance in the target region, the target pattern, and the target boundary. We computed these cues on the set of images. They form two approximately trivariate normal distributions for the target present and target absent categories. We then computed the decision boundary, error rate and $d'_b$ against varying conditions. Fig. \ref{fig:10_examples_vision}b shows the result for one condition, with a hyperboloidal boundary. The color-bar of the proportions of the contributions $\delta^{\text{rem}}_i$ to the overall $d'_b$ shows that the template provides most of the discriminability, followed by the edge, then the silhouette. These error rates, $d'_b$'s, and relative contributions to it by each feature can then be compared across conditions.

\subsection{Detecting camouflage}

We also applied this method in a study measuring performance in detecting camouflaged objects \cite{das2018understanding}. The major cue for detecting the object (fig. \ref{fig:10_examples_vision}c) is its edge, which we compute at scales of 2px, 4px and 8px. We extract two scalar features from the edge at each scale: the edge power captures its overall prominence, and the edge spectrum characterizes how this prominence is distributed along the boundary. We thus have 6 total features. Fig. \ref{fig:10_examples_vision}d shows the classification of these images using these 6 features, projected onto the Bayes decision variable (log likelihood ratio) $l$. The color-bars show that the proportion of contributions $\delta^{\text{rem}}_i$ to the overall discriminability are somewhat different from the individual $d'_i$ of each feature. We see that the first feature (bottom color of the bars), edge power at 2px, accounts for most of the detectability. In the reduced dimension of this plot, we can see that the absent distribution is quite normal, and present is nearly so. Consistently, in a normality test for classification with 100 trials, fig. \ref{fig:10_examples_vision}e, the hit fraction deviates only marginally from its normal prediction, so we accept the normal model here. Fig. \ref{fig:10_examples_vision}f shows classification using only the 2px features.  We use our dimension reduction technique to combine these two cues into the Bayes decision variable $l$ of this space, which we call simply the 2px edge cue. Classifying using this single variable, fig. \ref{fig:10_examples_vision}g, is the same as the 2d classification of fig. \ref{fig:10_examples_vision}f, and preserves the errors. We do the same merging at 4px and 8px, thus mapping 6 features to 3. Fig. \ref{fig:10_examples_vision}h shows the classification using these 3 merged cues. Due to the information in the two added scales, the classification has improved, and we see again that the 2px edge information accounts for almost all of the performance. The total number of classifier parameters used in this sequential classification is 28 (6 for each of the three 2d classifiers, then 10 when combining them in 3d). The classifier in full 6d has 28 parameters as well, yet it performs better since it can simultaneously optimize them all. Even so, merging features allows one to combine them in groups and sequences according to the problem structure, and visualize them.

\section{Conclusions}
In this paper, we presented our methods and open-source software for computing integrals and classification performance for normal distributions over a wide range of situations.

We began by describing how to integrate any multinormal distribution in a quadratic domain using the generalized chi-square method, then presented our ray-trace method to integrate in any domain, using examples from our software. We explained how this is synonymous with computing cdf's of quadratic and arbitrary functions of normal vectors, which can then be used to compute their pdf's and inverse cdf's as well.

We then described how to compute, given the parameters of multiple multinormals or labelled data from them, the classification error matrix, with optimal or sub-optimal classifiers, and the maximum (Bayes-optimal) discriminability index $d'_b$ between two normals. We showed that the common indices $d'_a$ and $d'_e$ underestimate $d'_b$, and that contrary to common use, $d'_e$ is often a better approximation than $d'_a$, even for two-interval tasks. We presented two methods to scale the discriminability of data, to say, match a model to a subject or observations.

We next described methods to merge and reduce dimensions for normal integration and classification problems without losing information. We presented tests for how reliably all the above methods, which assume normal distributions, can be used for other distributions. We followed this by demonstrating the speed and accuracy of the methods and software on different problems.

Finally, we illustrated all of the above methods on two visual detection research projects from our laboratory.

Although not developed here, the approach of our ray-trace integration method may carry over to other univariate and multivariate distributions. In the method, we spherically symmetrize the normal, find its distribution along any ray from the center, then add it over a grid of angles. This transforms all problem shapes to the canonical spherical form, then efficiently integrates outward from the center of the distribution. Some distributions, e.g. log-normal, can simply be transformed to a normal and then integrated with this method. For example, if $y \sim \text{lognormal}(\mu=1, \sigma=0.5)$, then we can compute, say, $p(\sin y >0) = p(\sin e^x>0)=0.65$ (where $x$ is normal), and all other quantities such as pdf's, cdf's, and inverse cdf's of its arbitrary functions (see toolbox example guide).

For other distributions, our general method is still useful if they are already spherically symmetric (i.e. spherical distributions), or can be made so (e.g. elliptical distributions), and the ray distribution through the sphere can be found. When they cannot be spherized, the ray distribution (if calculable) will depend on the orientation, just as the integration domain does. But once this additional dependency has been taken into account, integrating along rays from the center should still be the efficient method for distributions that fall off away from their center.

\section{Acknowledgements}
We thank Dr. Johannes Burge (University of Pennsylvania), Dr. R Calen Walshe (University of Texas at Austin), Kristoffer Frey (MIT), Dr. David Marvin Green, Dr. Helios De Rosario-Martínez (Instituto de Biomecánica de Valencia), Shizhuang Wang (Shanghai Jiao Tong University), and Florian Roth (TU Dresden) for discussions, improvements and contributions in the method, code and text. This work was supported by NIH grants EY11747 and EY024662.

\bibliographystyle{unsrt}
\bibliography{references}

@article{das2021method,
  title={A method to integrate and classify normal distributions},
  author={Das, Abhranil and Geisler, Wilson S},
  journal={Journal of Vision},
  volume={21},
  number={10},
  pages={1--1},
  year={2021},
  publisher={The Association for Research in Vision and Ophthalmology}
}

@book{trefethen2019approximation,
  title={Approximation theory and approximation practice},
  author={Trefethen, Lloyd N},
  volume={164},
  year={2019},
  publisher={Siam}
}

@article{das2024new,
  title={New methods to compute the generalized chi-square distribution},
  author={Das, Abhranil},
  journal={arXiv preprint arXiv:2404.05062},
  year={2024}
}

@book{duda2012pattern,
  title={Pattern classification},
  author={Duda, Richard O and Hart, Peter E and Stork, David G},
  year={2012},
  publisher={John Wiley \& Sons}
}

@article{das2018understanding,
  title={Understanding camouflage detection},
  author={Das, Abhranil and Geisler, Wilson},
  journal={Journal of Vision},
  volume={18},
  number={10},
  pages={549},
  year={2018},
  publisher={The Association for Research in Vision and Ophthalmology}
}

@book{green1966signal,
  title={Signal detection theory and psychophysics},
  author={Green, David Marvin and Swets, John A.},
  volume={1},
  year={1966},
  publisher={Wiley New York}
}

@article{saff1997distributing,
  title={Distributing many points on a sphere},
  author={Saff, Edward B and Kuijlaars, A BJ},
  journal={The mathematical intelligencer},
  volume={19},
  number={1},
  pages={5--11},
  year={1997},
  publisher={Springer}
}

@article{walshe2020detection,
  title={Detection of occluding targets in natural backgrounds},
  author={Walshe, R Calen and Geisler, Wilson S},
  journal={Journal of Vision},
  volume={20},
  number={13},
  pages={14},
  year={2020},
  publisher={The Association for Research in Vision and Ophthalmology}
}

@article{ruben1960probability,
  title={Probability content of regions under spherical normal distributions, I},
  author={Ruben, Harold},
  journal={The Annals of Mathematical Statistics},
  volume={31},
  number={3},
  pages={598--618},
  year={1960},
  publisher={JSTOR}
}

@article{ruben1962probability,
  title={Probability content of regions under spherical normal distributions, IV: The distribution of homogeneous and non-homogeneous quadratic functions of normal variables},
  author={Ruben, Harold},
  journal={The Annals of Mathematical Statistics},
  volume={33},
  number={2},
  pages={542--570},
  year={1962},
  publisher={JSTOR}
}

@article{davies1973numerical,
  title={Numerical inversion of a characteristic function},
  author={Davies, Robert B},
  journal={Biometrika},
  volume={60},
  number={2},
  pages={415--417},
  year={1973},
  publisher={Oxford University Press}
}

@book{genz2009computation,
  title={Computation of multivariate normal and t probabilities},
  author={Genz, Alan and Bretz, Frank},
  volume={195},
  year={2009},
  publisher={Springer Science \& Business Media}
}

@article{imhof1961computing,
  title={Computing the distribution of quadratic forms in normal variables},
  author={Imhof, Jean-Pierre},
  journal={Biometrika},
  volume={48},
  number={3/4},
  pages={419--426},
  year={1961},
  publisher={JSTOR}
}

@article{ng2000cs229,
  title={Generative Learning Algorithms},
  author={Ng, Andrew},
  journal={CS229 Lecture notes},
  volume={IV},
  year={2019}
}

@article{orucc2003weighted,
  title={Weighted linear cue combination with possibly correlated error},
  author={Oru{\c{c}}, Ipek and Maloney, Laurence T and Landy, Michael S},
  journal={Vision research},
  volume={43},
  number={23},
  pages={2451--2468},
  year={2003},
  publisher={Elsevier}
}

@book{wickens2002elementary,
  title={Elementary signal detection theory},
  author={Wickens, Thomas D},
  year={2002},
  publisher={Oxford University Press, USA}
}

@article{paranjpe1994selecting,
  title={Selecting variables for discrimination when covariance matrices are unequal},
  author={Paranjpe, SA and Gore, AP},
  journal={Statistics \& probability letters},
  volume={21},
  number={5},
  pages={417--419},
  year={1994},
  publisher={Elsevier}
}

@article{chaddha1968empirical,
  title={An empirical comparison of distance statistics for populations with unequal covariance matrices},
  author={Chaddha, RL and Marcus, LF},
  journal={Biometrics},
  pages={683--694},
  year={1968},
  publisher={JSTOR}
}

@article{simpson1973best,
  title={What is the best index of detectability?},
  author={Simpson, Adrian J and Fitter, Mike J},
  journal={Psychological Bulletin},
  volume={80},
  number={6},
  pages={481},
  year={1973},
  publisher={American Psychological Association}
}

@techreport{egan1962psychophysics,
  title={Psychophysics and signal detection.},
  author={Egan, James P and Clarke, Frank R},
  year={1962},
  institution={Indiana University Hearing and Communication Laboratory}
}

@article{green2020homily,
  title={A homily on signal detection theory},
  author={Green, David M},
  journal={The Journal of the Acoustical Society of America},
  volume={148},
  number={1},
  pages={222--225},
  year={2020},
  publisher={Acoustical Society of America}
}

@article{genz2002comparison,
  title={Comparison of methods for the computation of multivariate t probabilities},
  author={Genz, Alan and Bretz, Frank},
  journal={Journal of Computational and Graphical Statistics},
  volume={11},
  number={4},
  pages={950--971},
  year={2002},
  publisher={Taylor \& Francis}
}

@article{kessy2018optimal,
  title={Optimal whitening and decorrelation},
  author={Kessy, Agnan and Lewin, Alex and Strimmer, Korbinian},
  journal={The American Statistician},
  volume={72},
  number={4},
  pages={309--314},
  year={2018},
  publisher={Taylor \& Francis}
}
\end{document}